\documentclass[11pt]{article}

\usepackage[utf8]{inputenc}
\usepackage[T1]{fontenc}
\usepackage{lmodern}
\usepackage{microtype}

\usepackage[a4paper, top=2.5cm, bottom=2.5cm, left=2.8cm, right=2.8cm]{geometry}

\usepackage{amsmath}
\usepackage{amssymb}

\usepackage{booktabs}
\usepackage{longtable}
\usepackage{array}
\usepackage{multirow}
\usepackage{calc}

\usepackage{graphicx}
\usepackage{xcolor}
\usepackage{float}

\usepackage[
  colorlinks=true,
  hyperfootnotes=false,
  linkcolor=blue!70!black,
  citecolor=blue!70!black,
  urlcolor=blue!70!black,
  pdfauthor={Bo Zou, Chao Xu},
  pdftitle={QUIET: A Multi-Blank Cascaded Story Cloze Benchmark
            for LLM Creative Generation Capability},
]{hyperref}
\usepackage{bookmark}

\usepackage{natbib}
\usepackage{parskip}
\usepackage{setspace}
\usepackage{newunicodechar}
\newunicodechar{α}{\ensuremath{\alpha}}
\newunicodechar{β}{\ensuremath{\beta}}
\newunicodechar{γ}{\ensuremath{\gamma}}
\newunicodechar{δ}{\ensuremath{\delta}}
\newunicodechar{ε}{\ensuremath{\varepsilon}}
\newunicodechar{ζ}{\ensuremath{\zeta}}
\newunicodechar{η}{\ensuremath{\eta}}
\newunicodechar{θ}{\ensuremath{\theta}}
\newunicodechar{λ}{\ensuremath{\lambda}}
\newunicodechar{μ}{\ensuremath{\mu}}
\newunicodechar{ν}{\ensuremath{\nu}}
\newunicodechar{ξ}{\ensuremath{\xi}}
\newunicodechar{π}{\ensuremath{\pi}}
\newunicodechar{ρ}{\ensuremath{\rho}}
\newunicodechar{σ}{\ensuremath{\sigma}}
\newunicodechar{τ}{\ensuremath{\tau}}
\newunicodechar{φ}{\ensuremath{\varphi}}
\newunicodechar{χ}{\ensuremath{\chi}}
\newunicodechar{ψ}{\ensuremath{\psi}}
\newunicodechar{ω}{\ensuremath{\omega}}
\newunicodechar{Δ}{\ensuremath{\Delta}}
\newunicodechar{Σ}{\ensuremath{\Sigma}}
\newunicodechar{Π}{\ensuremath{\Pi}}
\newunicodechar{Ω}{\ensuremath{\Omega}}
\newunicodechar{Φ}{\ensuremath{\Phi}}
\newunicodechar{Ψ}{\ensuremath{\Psi}}
\newunicodechar{Θ}{\ensuremath{\Theta}}
\newunicodechar{Λ}{\ensuremath{\Lambda}}
\newunicodechar{Γ}{\ensuremath{\Gamma}}
\newunicodechar{≈}{\ensuremath{\approx}}
\newunicodechar{≤}{\ensuremath{\leq}}
\newunicodechar{≥}{\ensuremath{\geq}}
\newunicodechar{≠}{\ensuremath{\neq}}
\newunicodechar{≡}{\ensuremath{\equiv}}
\newunicodechar{∼}{\ensuremath{\sim}}
\newunicodechar{∞}{\ensuremath{\infty}}
\newunicodechar{∈}{\ensuremath{\in}}
\newunicodechar{∉}{\ensuremath{\notin}}
\newunicodechar{∀}{\ensuremath{\forall}}
\newunicodechar{∃}{\ensuremath{\exists}}
\newunicodechar{∅}{\ensuremath{\emptyset}}
\newunicodechar{→}{\ensuremath{\rightarrow}}
\newunicodechar{←}{\ensuremath{\leftarrow}}
\newunicodechar{↔}{\ensuremath{\leftrightarrow}}
\newunicodechar{⇒}{\ensuremath{\Rightarrow}}
\newunicodechar{⇐}{\ensuremath{\Leftarrow}}
\newunicodechar{⇔}{\ensuremath{\Leftrightarrow}}
\newunicodechar{×}{\ensuremath{\times}}
\newunicodechar{·}{\ensuremath{\cdot}}
\newunicodechar{±}{\ensuremath{\pm}}
\newunicodechar{∓}{\ensuremath{\mp}}
\newunicodechar{∪}{\ensuremath{\cup}}
\newunicodechar{∩}{\ensuremath{\cap}}
\newunicodechar{∫}{\ensuremath{\int}}
\newunicodechar{∑}{\ensuremath{\sum}}
\newunicodechar{∏}{\ensuremath{\prod}}
\newunicodechar{√}{\ensuremath{\sqrt{\,}}}
\newunicodechar{∇}{\ensuremath{\nabla}}
\newunicodechar{∂}{\ensuremath{\partial}}
\newunicodechar{∝}{\ensuremath{\propto}}
\newunicodechar{ℝ}{\ensuremath{\mathbb{R}}}
\newunicodechar{ℕ}{\ensuremath{\mathbb{N}}}
\newunicodechar{ℤ}{\ensuremath{\mathbb{Z}}}
\newunicodechar{ℚ}{\ensuremath{\mathbb{Q}}}
\newunicodechar{①}{(1)}
\newunicodechar{②}{(2)}
\newunicodechar{③}{(3)}
\newunicodechar{④}{(4)}
\newunicodechar{⑤}{(5)}
\newunicodechar{⑥}{(6)}
\newunicodechar{⑦}{(7)}
\newunicodechar{⑧}{(8)}
\newunicodechar{⑨}{(9)}
\newunicodechar{⑩}{(10)}
\newunicodechar{✓}{\ensuremath{\checkmark}}
\newunicodechar{✗}{\ensuremath{\times}}
\newunicodechar{∧}{\ensuremath{\wedge}}
\newunicodechar{∨}{\ensuremath{\vee}}
\newunicodechar{¬}{\ensuremath{\neg}}
\newunicodechar{★}{$\bigstar$}
\newunicodechar{☆}{$\star$}
\newunicodechar{●}{$\bullet$}
\newunicodechar{○}{$\circ$}
\newunicodechar{■}{$\blacksquare$}
\newunicodechar{□}{$\square$}

\title{\textbf{QUIET: A Multi-Blank Cascaded Story Cloze Benchmark\\
for LLM Creative Generation Capability}\\[0.4em]
\large\itshape Quality Understanding via Interlocked Evaluation Testing}
\author{Bo Zou \and Chao Xu}
\date{2026}

\newcounter{none}
\providecommand{\tightlist}{\setlength{\itemsep}{0pt}\setlength{\parskip}{0pt}}

\begin{document}
\maketitle

\begin{abstract}

Large language models (LLMs) face a dual challenge in creative capability evaluation: existing benchmarks (e.g., Story Cloze Test, HellaSwag) measure models' discriminative ability over narrative continuation using multiple-choice recognition paradigms, rather than directly measuring creative generation capability; rubric-based scoring and LLM-as-Judge methods rely on subjective dimension assessment or natural language model outputs, and cannot provide objective, automated scoring mechanisms.

This paper proposes \textbf{QUIET} (Quality Understanding via Interlocked Evaluation Testing), a diagnostic benchmark for LLM creative capability based on multi-blank cascaded story cloze. QUIET sets \(N\) blanks (10--20) in a story with complete structure, with each blank accompanied by an explicit content constraint, and cascade dependency relationships between blanks --- the content filled into earlier blanks constrains the feasible solution space for later blanks. The evaluated model (or human participants) fills all blanks in open-ended generation mode; the results are scored by an information-theoretic automated scoring protocol without human grading.

The scoring protocol directly operationalizes the ``calibrated surprise'' theoretical framework (Zou \& Xu, 2026a). For each blank \(k\), a composite score is computed:

\[\text{score}(x_k) = \text{satisfy}(x_k, c_k) \times \left(1 + \lambda \cdot \text{surprise}(x_k)\right), \quad \lambda = 1.0\]

where \(\text{satisfy}\) measures how well the blank filling satisfies the content constraint (objective logical reasoning judgment, not subjective aesthetic scoring), and \(\text{surprise}\) measures the degree of surprise given that the constraint is satisfied. Creative answers that do not satisfy the constraint score zero; answers that satisfy the constraint but are mediocre score low; answers that satisfy the constraint and are surprising score high. This scoring structure belongs to a mathematically distinct category from rubric's linear weighted sum \(Q = \sum w_i s_i\) --- QUIET's per-blank constraint satisfaction check is an objective logical reasoning judgment at a specific narrative position, whereas rubric involves subjective dimension scoring spanning the entire work.

This paper further formalizes the information-theoretic model of multi-blank cascades --- the longitudinal unfolding of the chain rule \(I(X_1,...,X_N; Y) = \sum_k I(X_k; Y \mid X_{<k})\) along the narrative time axis --- and introduces ``retrospective sensitivity'' \(\rho(X_t)\) to characterize the distribution of narrative constraints along the time axis. We demonstrate how multi-blank full-structure design naturally avoids the retrospective evaluation dilemma faced by single-blank paradigms.

Empirical validation on 12 commercial LLMs with significant generational differences shows that the QUIET composite score effectively distinguishes previous-generation baseline models from modern flagship models: GPT-3.5-turbo (6.44/14) significantly trails the other 11 models (7.46--8.69). Within the modern model cluster, Gemini-3.1-Pro (8.69) and Gemini-2.5-pro (8.44) rank first and second, with Claude-Opus-4.6 (8.38) closely following GLM-5.1-thinking (8.40) in third-fourth place (gap of 0.02). Spearman sensitivity analysis shows that the core conclusion (GPT-3.5 at the bottom, Gemini family at the top) remains robust across 14 aggregation formula combinations. This paper also transparently reports a known limitation of the LLM-as-Judge scoring system --- Krippendorff α ≈ 0.27 moderate inter-rater agreement. We additionally run a scoring scale granularity ablation experiment, applying both fine-grained 6-point scoring (main results, α=0.270) and coarse 3-tier 0/0.5/1.0 scoring (ablation control) to the same answer set; α remains almost unchanged. \textbf{This indicates that the low inter-rater agreement of LLM-as-Judge in creative constraint evaluation stems from systematic judgment-criterion drift between judge models, not from scoring scale granularity} --- this negative result has diagnostic value, suggesting that future improvements should prioritize introducing human-anchored calibration rather than continuing to refine the scale at the prompt level.

\textbf{Human behavioral distribution evidence prior to the automated scoring protocol.} QUIET's theoretical framework was not reverse-engineered from LLM data. Before the automated scoring protocol was designed, the same cloze test was distributed in March 2026 to 135 literary enthusiasts via invitation-only, with no scoring prompts or human intervention. The participant group spontaneously stratified into three behavioral patterns: ``mid-task dropout'' (N=100, 74.07\%), ``submitted-but-mediocre'' (N=31, 22.96\%), and ``shortlisted'' (N=4, 2.96\%) --- precisely corresponding to the three theoretical quadrants of the chain rule \(I(X_k; Y \mid X_{<k}) \to 0\), ``accurate without surprise,'' and ``calibrated surprise'' (§5.6). This \textbf{human behavioral distribution that emerged prior to framework construction} constitutes independent prior empirical evidence for QUIET's theoretical framework; it also forms a critical contrast with LLM experiments --- all 12 models completed 36 blank fillings, with no model exhibiting the human ``giving up under difficulty'' cascaded perception behavior, revealing a meta-cognitive evaluation dimension long inaccessible to single-blank benchmarks.

To our knowledge, QUIET is the first benchmark to provide automated scoring of LLM creative capability under the open-ended generation (rather than multiple-choice discrimination) paradigm. This paper positions QUIET as a stepping stone in this direction: the issues exposed by the current implementation --- low inter-rater agreement and limited discrimination within the modern model cluster --- reflect \textbf{common challenges in the early stage} of this research direction, not structural defects in the QUIET framework itself. §6.4 provides a clear improvement path for introducing human-anchored calibration and atomic checks.

\end{abstract}

\section{Introduction}\label{introduction}

\subsection{The Problem: A Triple Dilemma in LLM Creative Evaluation}\label{the-problem-a-triple-dilemma-in-llm-creative-evaluation}

LLM creative capability evaluation currently relies on three pathways, each with critical flaws.

\textbf{Pathway 1: Multiple-choice recognition benchmarks.} Story Cloze Test (Mostafazadeh et al., 2016), HellaSwag (Zellers et al., 2019), and similar benchmarks adopt a multiple-choice paradigm --- given narrative context and several candidate continuations, models are evaluated on discriminative ability via logprobs or accuracy. These benchmarks have two layers of limitations. First, \textbf{paradigm-level limitation}: multiple-choice recognition measures the ability to ``identify the best continuation from given options'' (discriminative capability), not ``autonomously generate high-quality continuation'' (creative generation capability) --- just as a skilled literary critic does not necessarily write good fiction, a model that can select the best continuation may not generate text of equivalent quality. Even with improvements within the multiple-choice paradigm (such as finer blank granularity or dimension-specific distractors), as long as the task form remains ``choose from given options,'' what is measured is fundamentally discriminative capability, not creative generation capability. Second, \textbf{structural-level limitation}: these benchmarks all use single-blank design --- only one decision point in the narrative --- unable to measure a creator's ability to maintain consistency and build cascaded logic across multiple decision points.

\textbf{Pathway 2: Rubric scoring.} Rubric methods decompose quality a priori into several sub-dimensions (e.g., ``character development,'' ``plot logic,'' ``writing style''), scoring each sub-dimension independently and computing a weighted sum, implicitly assuming linear additivity \(\text{Quality} = \sum_i w_i \cdot s_i\). This assumption is mathematically equivalent to discarding all interaction effects between dimensions (§6.1 details this). Ye et al.~(ICLR 2024) empirically show that inter-rater agreement for rubric evaluation in creative quality is far lower than for factual tasks; Messick (1994) argues from validity theory for the structural risks of decomposing complex capabilities into independent sub-items.

\textbf{Pathway 3: LLM-as-Judge.} Using LLMs as judges requires models to output natural language scoring rationale. This measures the model's evaluative expression capability rather than its internal probability distribution's actual sensitivity to creative constraints --- a model can output reasonable scoring rationale while its internal probability distribution is completely insensitive to the corresponding constraints.

All three pathways share one fundamental absence: \textbf{none can directly measure LLM creative generation capability in an objective, automated, and reproducible manner.}

\subsection{Our Proposal: QUIET}\label{our-proposal-quiet}

This paper proposes \textbf{QUIET} (Quality Understanding via Interlocked Evaluation Testing), a diagnostic benchmark for LLM creative capability. Its design includes the following core innovations.

\textbf{Multi-blank cascaded story cloze.} \(N\) blanks (10--20) are set in a story with complete structure. A blank can be a word, a sentence, or a paragraph. Blanks have explicit cascade dependency relationships --- the content filled into earlier blanks directly constrains the feasible solution space for later blanks. This design forces the evaluated party to make each local creative decision within a global narrative perspective, simulating the cascaded thinking characteristic of real creative processes.

\textbf{Open-ended generation rather than multiple-choice recognition.} QUIET provides no candidate options; the evaluated model must generate blank fillings itself. This design directly upgrades the evaluation target from ``identifying good creative decisions'' (discriminative capability) to ``making good creative decisions'' (creative generation capability).

\textbf{Information-theoretic automated scoring.} Each blank is accompanied by an explicit content constraint (e.g., ``this blank must explain the reason for the protagonist's abnormality''). Scoring is performed automatically by machine without human grading. The scoring protocol directly operationalizes the core claim of the ``calibrated surprise'' paper (Zou \& Xu, 2026a): constraint satisfaction (accurate) × surprise (surprising) = composite score.

\textbf{Structural distinction from rubric.} QUIET's per-blank constraint satisfaction scoring is mathematically distinct from rubric. Rubric assigns scores on several abstract dimensions across the entire work --- these are subjective aesthetic judgments spanning the entire piece. QUIET checks a specific content constraint for each blank (e.g., ``whether the text explains the reason for X'') --- this is objective logical reasoning at a specific position, closer to exam grading than literary review. §6.1 elaborates on this distinction.

\subsection{Contribution Statement}\label{contribution-statement}

The main contributions of this paper are as follows:

\begin{enumerate}
\def\labelenumi{\arabic{enumi}.}
\tightlist
\item
  We propose the first LLM creative capability benchmark based on multi-blank cascaded cloze and open-ended generation, upgrading the evaluation paradigm from discriminative capability measurement to creative generation measurement.
\item
  We design an automated scoring protocol based on constraint satisfaction × surprise --- directly operationalizing the ``calibrated surprise'' information-theoretic framework, without human grading, and provably distinct from rubric in mathematical structure.
\item
  We demonstrate the design goal of this scoring mechanism: to compress the aesthetic dependency of creative generation evaluation into two basic capabilities --- ``per-blank constraint logical judgment + geometric distance computation'' --- where constraint satisfaction judgment is a logical reasoning task (NLI) and surprise computation is a geometric distance estimation, neither requiring the scoring model itself to possess high-order creative aesthetics. It should be noted that the current implementation still uses holistic LLM-as-Judge for the satisfy component (6-point scale + knockout clause, α=0.270, see §5.1), retaining some subjective space; the follow-up scoring scheme targeting human-anchored calibration (§6.4) aims to further eliminate this residual subjectivity. Even so, QUIET is already the first benchmark to explicitly decompose creative generation evaluation into two independently developable components --- ``logical judgment × geometric distance.''
\item
  We formalize the information-theoretic model of multi-blank cascaded constraints: the longitudinal unfolding of the chain rule \(I(X_1,...,X_N; Y) = \sum_k I(X_k; Y \mid X_{<k})\) along the narrative time axis.
\item
  We introduce the formal definition of retrospective sensitivity \(\rho(X_t)\), characterizing the distribution of narrative constraints along the time axis, and demonstrate how multi-blank full-structure design naturally avoids the retrospective evaluation problem faced by single-blank paradigms.
\item
  We report qualitative findings from a pilot human cloze study (§5.6) --- the same test was distributed via invitation-only in an online literary community, where participants spontaneously stratified into three behavioral patterns --- ``mid-task dropout,'' ``submitted-but-mediocre,'' and ``shortlisted'' --- without any experimenter intervention, corresponding respectively to the extreme form of the chain rule when \(I(X_k; Y \mid X_{<k}) \to 0\), the ``accurate without surprise'' quadrant of the CQA framework, and the ``calibrated surprise'' quadrant. This three-tier distribution was observed before the construction of QUIET's theoretical framework and automated scoring protocol, constituting independent prior empirical evidence for the §2 theoretical framework; it also forms a critical contrast with LLM experimental results --- all 12 models completed all 36 blank fillings, with no model exhibiting the ``giving up under difficulty'' cascaded perception behavior of human participants.
\item
  We empirically validate QUIET's discriminative power and scoring protocol effectiveness on 12 commercial LLMs with significant generational variation (including 1 previous-generation baseline model GPT-3.5-turbo released in 2022, and 11 modern flagship/main commercial models released in 2025--2026), reporting three-dimensional results: constraint satisfaction, surprise, and QUIET composite score. Experimental results are consistent with theoretical predictions in the main trend, while identifying two improvable aspects of the current scoring protocol: inter-rater agreement (α=0.270, with scoring scale granularity eliminated as a factor) and discrimination within the modern model cluster.
\end{enumerate}

\textbf{A note on positioning.} To our knowledge, QUIET is the first benchmark to provide automated scoring of LLM creative capability under the open-ended generation paradigm. This paper positions it as a \emph{stepping stone} in this direction: the low inter-rater agreement and limited discrimination within the modern model cluster exposed by the current implementation reflect \textbf{common challenges in the early stage} of this research direction, not structural defects of the QUIET framework itself. §6.4 provides a clear improvement path for introducing human-anchored calibration and atomic checks. This paper invites reviewers to evaluate this work on the scale of ``the first benchmark in a new direction'' --- prioritizing the soundness of the theoretical framework, the originality of the design, and the differentiation from existing benchmarks --- rather than the scale of ``a mature and finalized benchmark.''

\section{Theoretical Foundation}\label{theoretical-foundation}

This section builds the theoretical framework of this benchmark starting from the core results of ``Calibrated Surprise'' (Zou \& Xu, 2026a). §2.1 introduces the information-theoretic tools; §2.2 establishes the chain rule model for multi-blank cascades; §2.3 introduces retrospective sensitivity and demonstrates how multi-blank design naturally avoids retrospective evaluation problems; §2.4 demonstrates the structural differences between QUIET scoring and rubric.

\subsection{Preliminary Definitions}\label{preliminary-definitions}

The theoretical tools in this paper come from the framework established in ``Calibrated Surprise'' (Zou \& Xu, 2026a). That framework measures creative quality as mutual information \(I(X;Y) = H(X) - H(X|Y)\), where \(X\) is the creative choice and \(Y\) is the intersection of all-dimension real-world constraints (psychological authenticity, causal logic, world consistency, tonal fit, thematic depth, etc.).

A key distinction: high \(I(X;Y)\) requires two conditions simultaneously ---

\begin{itemize}
\tightlist
\item
  \textbf{Accurate} (calibrated): \(H(X|Y) \approx 0\), meaning the creative choice precisely satisfies the constraint, with an extremely narrow feasible solution space.
\item
  \textbf{Surprising}: \(H(X)\) is high, meaning that without knowing the constraints, this choice is not a mediocre ``default option.''
\end{itemize}

Accuracy without surprise = mediocre correctness (the standard answer anyone could write); surprise without accuracy = absurd nonsense; both together = \textbf{calibrated surprise} --- the hallmark of higher-order creative work. For the complete argument of this framework, see (Zou \& Xu, 2026a).

\subsection{The Chain Rule for Multi-Blank Cascades}\label{the-chain-rule-for-multi-blank-cascades}

§2.1 defines creative quality as mutual information for a single decision. But real creative work is a sequence of decisions: every creative choice in a story builds on all previous choices. The multi-blank cascaded design of this benchmark is the direct operationalization of this sequential structure.

Consider \(N\) blanks \(X_1, X_2, ..., X_N\) in a story, with the story skeleton providing constraints \(Y\). The \textbf{total creative information} demonstrated by filling all blanks can be decomposed by the chain rule:

\[I(X_1, X_2, ..., X_N;\, Y) = \sum_{k=1}^{N} I(X_k;\, Y \mid X_{<k})\]

where \(X_{<k} = (X_1, ..., X_{k-1})\) represents all previously filled blanks. Each term \(I(X_k; Y \mid X_{<k})\) measures: \textbf{given all previously filled blanks, the additional creative information contributed by the \(k\)-th blank filling.}

\textbf{Mathematical meaning of cascaded constraints.} The condition \(X_{<k}\) in the chain rule encodes cascade dependency. Consider this example:

\begin{itemize}
\tightlist
\item
  Blank 1: ``My brain is not normal; its main symptom is {[}blank{]}'' --- constraint \(c_1\): must explain the nature or cause of the brain abnormality
\item
  Blank 3: ``Because of this, at school I {[}blank{]}'' --- constraint \(c_3\): must present the specific consequences in a school setting caused by the brain abnormality described in Blank 1
\end{itemize}

If the evaluated party fills Blank 1 with ``unable to distinguish between men and women,'' the constraint space of Blank 3 is locked --- only school consequences logically consistent with ``unable to distinguish between men and women'' are acceptable answers. Mathematically, the computation of \(I(X_3; Y \mid X_1, X_2)\) is conditioned on the specific content of Blank 1, tightly coupling the ``accurate'' requirement of Blank 3 to Blank 1.

\textbf{Formal distinction between ``flat and mundane'' and ``calibrated surprise.''} Two extreme cases:

\begin{itemize}
\tightlist
\item
  \textbf{Flat and mundane}: every \(I(X_k; Y \mid X_{<k})\) is low --- the evaluated party's every blank filling satisfies the constraint but is the most obvious default option. Total score low.
\item
  \textbf{Calibrated surprise}: every \(I(X_k; Y \mid X_{<k})\) is high --- the evaluated party's every blank filling simultaneously satisfies the constraint and breaks expectations. Total score high.
\end{itemize}

This distinction directly corresponds to phenomena observed in the 2026 story cloze activity: most participants (including literary enthusiasts) could achieve ``every blank is accurate but unsurprising,'' while only very few could simultaneously achieve accurate and surprising answers across multiple blanks.

\subsection{Retrospective Sensitivity and Multi-Blank Full-Structure Design}\label{retrospective-sensitivity-and-multi-blank-full-structure-design}

The mutual information framework in §2.1 has an implicit premise: the evaluator must have complete information about constraint \(Y\). But in narrative creative work, the information in \(Y\) is non-uniformly distributed along the time axis --- some constraints come from the preceding context, some from later text or the ending. This section formalizes this temporal dimension and demonstrates how QUIET's multi-blank full-structure design naturally avoids the evaluation dilemma it creates.

\textbf{The problem.} For a creative decision \(X_t\) at narrative position \(t\), the full-dimension constraint \(Y\) can be decomposed temporally as:

\[Y = Y_{\leq t} \cup Y_{>t}\]

where \(Y_{\leq t}\) is the forward constraint and \(Y_{>t}\) is the backward/ending constraint. A real phenomenon in literary creation is that some decisions can be fully evaluated from the preceding context alone, while others can only be evaluated after the later text is revealed. Sternberg (2003) calls the latter ``retrospective reinterpretation'' --- readers return to earlier passages after the ending is revealed and reassess the significance of early decisions. van Dijk \& Kintsch (1983) addressed a related cognitive mechanism via the distinction between local coherence and global coherence.

\textbf{Formal definition.} Define the \textbf{retrospective sensitivity} of decision \(X_t\):

\[\rho(X_t) = \frac{I(X_t;\, Y_{>t} \mid Y_{\leq t})}{I(X_t;\, Y)}\]

\(\rho \in [0,1]\) forms a continuous spectrum: \(\rho \approx 0\) for retrospection-invariant decisions (the preceding context already contains sufficient constraint information), and \(\rho \gg 0\) for retrospection-sensitive decisions (key constraints are in the later text).

\textbf{The dilemma of single-blank paradigms.} Existing benchmarks (HellaSwag, Story Cloze, etc.) and the logprobs multiple-choice scheme developed in our team's early work all use single-blank paradigms --- only providing the preceding context \(Y_{\leq t}\), not the later context \(Y_{>t}\). When \(\rho \gg 0\) (e.g., foreshadowing-twist structures), evaluation based solely on preceding context will be systematically inaccurate.

\textbf{The natural advantage of multi-blank full-structure design.} QUIET's design naturally avoids this dilemma. The reason is that the story skeleton provides a complete structure --- including all text before, between, and after the blanks. The evaluated party can see the context before and after each blank when filling; and when scoring, the constraint condition \(c_k\) for each blank already encodes global narrative information (the task authors consulted the complete story structure when designing \(c_k\)). Therefore, QUIET's evaluation protocol is functionally equivalent to providing both \(Y_{\leq t}\) and \(Y_{>t}\), and \(\rho\) no longer constitutes a limiting factor for the measurement scope.

It should be noted that \(\rho\) retains independent value as a theoretical tool --- it is the first formalization of the temporal distribution of narrative constraints and can be used to analyze the implicit scope assumptions of different benchmarks (§6.2).

\subsection{Structural Differences Between QUIET Scoring and Rubric}\label{structural-differences-between-quiet-scoring-and-rubric}

QUIET's per-blank constraint satisfaction scoring differs from rubric in two irreducible structural ways.

\textbf{Different mathematical structure.} The rubric scoring function is \(Q = \sum_i w_i \cdot s_i\), where \(s_i\) is a dimension score spanning the entire work (e.g., ``plot logic = 4 points,'' ``character development = 3 points''). QUIET's scoring function is \(Q = \sum_k \text{satisfy}(x_k, c_k) \times (1 + \lambda \cdot \text{surprise}(x_k))\) (λ=1.0), where each \(\text{satisfy}(x_k, c_k)\) is a satisfaction check for a specific content constraint \(c_k\) at a specific narrative position \(k\). The key difference: rubric's \(s_i\) is a subjective aesthetic judgment of an abstract dimension across the entire work; QUIET's \(\text{satisfy}(x_k, c_k)\) is an objective logical reasoning judgment at a specific position --- ``whether this text explains the reason for X'' and ``how many points does the plot logic of this story deserve'' are entirely different cognitive operations.

\textbf{Bandwidth gap.} For a rubric with 5 dimensions × 5-point scale, the information capacity per evaluation is approximately \(5 \times \log_2 5 \approx 11.6\) bits. QUIET's per-blank scoring produces a continuous score for each of \(N = 15\) blanks, with information bandwidth far exceeding rubric's discrete scale.

\section{Test Set Construction}\label{test-set-construction}

\subsection{Story Selection and Blank Placement}\label{story-selection-and-blank-placement}

\textbf{Story source.} QUIET's story materials come from original mystery stories created by the team's core creative experts. All story materials have not been published on public internet channels, eliminating the risk of contamination from evaluated models' pretraining data.

\textbf{Blank placement principles.} Each story has 10--20 blanks, with blank granularity determined by narrative needs:

\begin{itemize}
\tightlist
\item
  \textbf{Word-level blanks}: key nouns, adjectives, or verbs (e.g., character emotion words, key scene objects)
\item
  \textbf{Sentence-level blanks}: a descriptive sentence or dialogue (e.g., a character's key line, a transitional sentence for scene change)
\item
  \textbf{Paragraph-level blanks}: a narrative passage or scene setting (e.g., a character's complete response to a crisis, the development of a key event)
\end{itemize}

Blank placement focuses on \textbf{key decision points} in the narrative --- positions where creators must make non-trivial choices: character motivation revelation, event turning points, detail selection, emotional description, causal inference, etc.

\subsection{Constraint Design}\label{constraint-design}

One of QUIET's core methodological contributions is the design of \textbf{integrated constraint conditions}. Unlike per-blank independent scoring, QUIET organizes several consecutive blanks into a \textbf{blank group}, with the entire blank group sharing a comprehensive content constraint condition \(c_g\). Constraint conditions are written by task authors (creative experts) when designing the test, functioning like exam ``scoring criteria'' or ``model answer explanations.''

\textbf{Why blank groups rather than per-blank scoring?} Creative decisions are not isolated points but a set of mutually coupled choices. In the actual test, a narrative passage (1--2 paragraphs) contains multiple blanks (e.g., blanks 01--06); these blanks jointly form a complete creative decision unit --- their fillings must be mutually coordinated and collectively satisfy a holistic constraint condition. Per-blank independent scoring would artificially sever this coupling, just as per-dimension rubric scoring loses dimension interaction effects (§6.1). Blank group design sits between ``per-blank'' and ``whole-work'' in scoring granularity, preserving the interaction information among blanks within a group while maintaining global coherence through cascade dependencies between groups.

Constraint design principles:

\begin{enumerate}
\def\labelenumi{\arabic{enumi}.}
\tightlist
\item
  \textbf{Content-level, not dimension-level.} Constraint conditions are specific content requirements for that story passage (e.g., ``must establish the specific nature of the protagonist's brain abnormality, and this nature must cause pain; must establish a job related to this nature that ordinary people cannot complete''), not abstract dimension scoring criteria.
\item
  \textbf{Objectively determinable.} Whether the constraint is satisfied can be determined by logical reasoning --- it is a factual judgment, not an aesthetic preference.
\item
  \textbf{Multi-blank coupling.} The fillings in the blanks within a group must mutually coordinate to collectively satisfy the constraint condition --- a single blank alone may not be judgeable, but their combination can be evaluated.
\item
  \textbf{Marked cascade dependencies.} Constraint conditions must mark the cascade relationships between this blank group and other blank groups (see §3.3).
\end{enumerate}

Constraint condition example (from the actual test):

{\def\LTcaptype{none} 
\begin{longtable}[]{@{}
  >{\raggedright\arraybackslash}p{(\linewidth - 4\tabcolsep) * \real{0.3333}}
  >{\raggedright\arraybackslash}p{(\linewidth - 4\tabcolsep) * \real{0.3333}}
  >{\raggedright\arraybackslash}p{(\linewidth - 4\tabcolsep) * \real{0.3333}}@{}}
\toprule\noalign{}
\begin{minipage}[b]{\linewidth}\raggedright
Blank Group
\end{minipage} & \begin{minipage}[b]{\linewidth}\raggedright
Blanks Included
\end{minipage} & \begin{minipage}[b]{\linewidth}\raggedright
Constraint Condition \(c_g\) (multiple items)
\end{minipage} \\
\midrule\noalign{}
\endhead
\bottomrule\noalign{}
\endlastfoot
Group 1 & Blanks 01--07 & ① Must establish a brain abnormality trait that causes pain ② The protagonist's job must be causally related to the trait ③ The job must be something ordinary people cannot complete ④ The job must produce or encounter something ordinary people cannot tolerate ⑤ The protagonist must carry an item A that helps with work or maintains normality \\
Group 2 & Blanks 08--12 & ① Must establish a reasonable reason for the protagonist not using a phone ② Item A's size must fit in an outer coat pocket ③ The protagonist also loses another item B, with a logical connection to A \\
Group 3 & Blanks 13--17 & ① Item A can be temporarily made but requires special materials ② Item A has a backup, which has an ``archival'' nature and can only be used temporarily \\
\end{longtable}
}

\subsection{Cascade Constraint Graph}\label{cascade-constraint-graph}

The dependency relationships between blank groups can be represented as a directed acyclic graph (DAG). Each node in the graph is a blank group; a directed edge \((g_i \to g_j)\) indicates that the content filled into blank group \(g_i\) constrains the feasible solution space of blank group \(g_j\).

Using the actual test as an example, the mystery story test set used in this paper contains 7 blank groups (36 blanks), with the following cascade constraint graph:

\begin{figure}[H]
  \centering
  \includegraphics[width=0.95\linewidth]{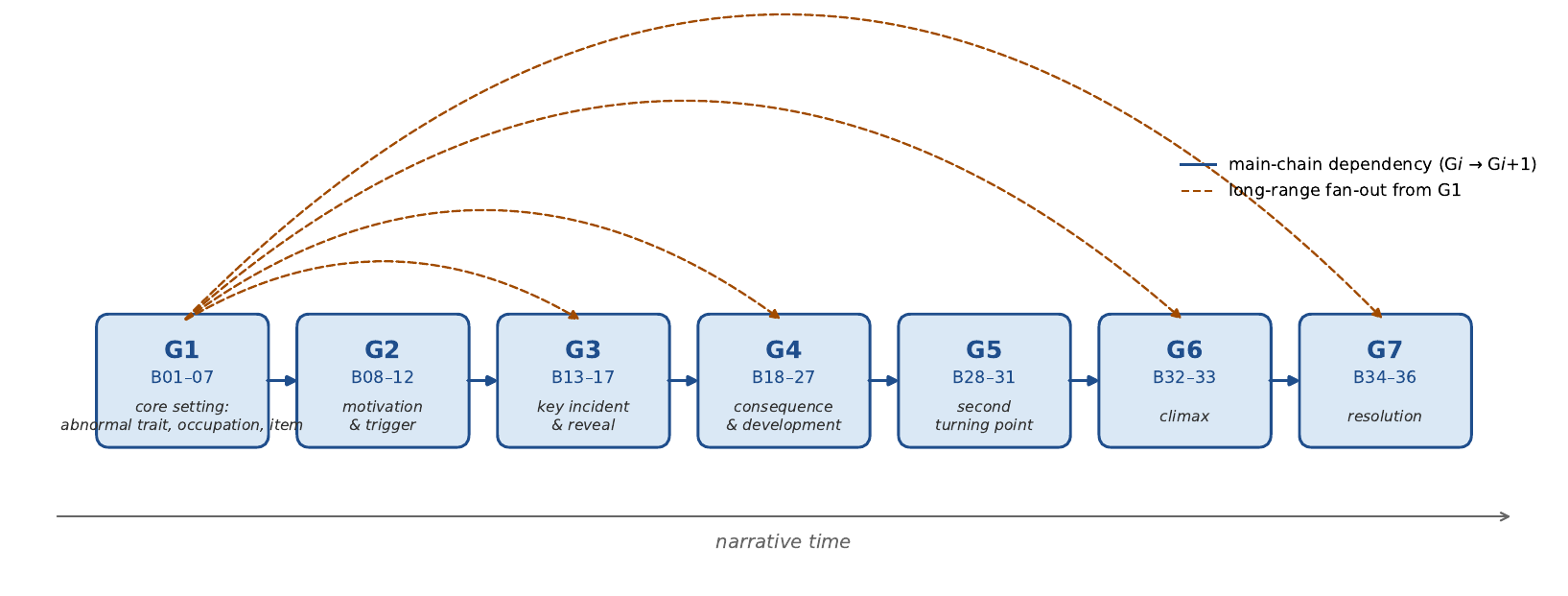}
  \caption{Cascade constraint DAG of the mystery-story test set (7 blank groups,
  36 blanks). Solid edges denote main-chain dependencies; dashed edges denote
  long-range fan-out constraints originating from G1 (which fixes the
  protagonist's abnormal trait, occupation, and key item).}
  \label{fig:cascade-dag}
\end{figure}

Structural characteristics of the cascade constraint graph:

\begin{enumerate}
\def\labelenumi{\arabic{enumi}.}
\tightlist
\item
  \textbf{The main trunk is a chain dependency.} G1 → G2 → G3 → G4 → G5 → G6 → G7 forms the main chain --- each group's filling must be logically consistent with the previous group.
\item
  \textbf{G1 is the source node with fan-out radiation.} G1 establishes the core parameters of the entire story (the protagonist's abnormal trait, occupation, key item A); all subsequent groups' constraint spaces are locked by G1. This makes G1's filling choices have the greatest impact on global scores --- in the chain rule \(I(X_k; Y \mid X_{<k})\), the marginal information contribution at \(k=1\) is naturally highest.
\item
  \textbf{Constraint tightening is monotonic.} As the story progresses, the accumulated constraints from earlier become more numerous, and the feasible solution space of later groups monotonically narrows --- this is the information-theoretic explanation for why participants in the pilot study reported ``it gets harder and harder, the earlier settings block all later paths.''
\end{enumerate}

\subsection{Example: Blank Group 1 (Blanks 01--07)}\label{example-blank-group-1-blanks-0107}

Below is the complete structure of the first blank group in the actual test.

\textbf{Story skeleton (with blanks):}

\begin{quote}
My brain is not quite normal; one of its main symptoms is (Blank 01). As an adult, this problem grew worse and worse, until now I (Blank 02) every day, which is simply unbearable. Of course, it is precisely because of (Blank 03) that I am able to do (Blank 04), this job that ordinary people cannot complete --- after all, only someone like me who (Blank 05) can endure the constant (Blank 06) produced by the work, and I can't really complain. To avoid trouble, I have to (Blank 07) every day and carry it with me at all times; that could be said to be my lifeline.
\end{quote}

\textbf{Integrated constraint conditions (5 items):}

\begin{enumerate}
\def\labelenumi{\arabic{enumi}.}
\tightlist
\item
  Must establish a brain abnormality trait that causes pain
\item
  The protagonist's job must be causally related to the brain abnormality trait
\item
  The job must be something ordinary people cannot complete, and must be difficult
\item
  The job must produce or encounter something ordinary people cannot tolerate
\item
  The protagonist must carry item A, which helps complete the work or maintain normality
\end{enumerate}

Note: all 7 blanks must \textbf{collectively} satisfy the above 5 constraints --- it is impossible to judge any single blank in isolation, but the combination of all 7 must form a self-consistent whole. This is the core characteristic of ``blank group integrated scoring.''

\textbf{Cascade effect illustration:} This group is the source node (no preceding dependencies), but its fillings will lock the constraint spaces of all 6 subsequent groups. For example, if Blank 01 chooses ``unable to distinguish between men and women,'' then Blank 03 (occupation) must be a job requiring this trait, and Blank 06 (item carried) must relate to this trait or occupation --- all constraint spaces of Groups 2--7 are locked by this choice.

\subsection{Test Set Size and Statistics}\label{test-set-size-and-statistics}

\begin{itemize}
\tightlist
\item
  \textbf{Scale}: the initial version contains 1 mystery story × 36 blanks = 36 measurement points, organized into 7 blank groups
\item
  \textbf{Blank group distribution}: G1 (7 blanks) → G2 (5 blanks) → G3 (5 blanks) → G4 (10 blanks) → G5 (4 blanks) → G6 (2 blanks) → G7 (3 blanks)
\item
  \textbf{Number of constraint conditions}: 7 blank groups with a total of 20 constraint conditions
\item
  \textbf{Language}: Chinese
\item
  \textbf{Evaluated subjects}: 12 commercial LLMs with significant generational differences + human participants from online recruitment
\item
  \textbf{Data isolation}: story materials have not been published on public internet channels, eliminating pretraining contamination risk
\end{itemize}

\section{Scoring Protocol}\label{scoring-protocol}

QUIET's scoring protocol consists of three tiers: constraint satisfaction (§4.1), surprise (§4.2), composite scoring (§4.3), and cascade consistency scoring for handling cascade relationships (§4.4). All scoring is performed automatically by machine without human grading.

\subsection{Tier 1: Constraint Satisfaction}\label{tier-1-constraint-satisfaction}

For each blank \(k\), the degree to which the filling \(x_k\) satisfies constraint condition \(c_k\) is computed.

\textbf{Scoring method: Natural Language Inference (NLI) style judgment.} Constraint satisfaction checking is modeled as an NLI task:

\begin{itemize}
\tightlist
\item
  \textbf{Premise}: story context + blank filling \(x_k\)
\item
  \textbf{Hypothesis}: natural language description of constraint condition \(c_k\) (e.g., ``the text explains the reason for the protagonist's brain abnormality'')
\item
  \textbf{Judgment}: entailment / neutral / contradiction
\end{itemize}

\[\text{satisfy}(x_k, c_k) = P_{\text{NLI}}(\text{entailment} \mid \text{context} + x_k,\; c_k)\]

This score can take two forms:
- \textbf{Hard judgment (0/1)}: 1 when \(P(\text{entailment}) > \theta\), 0 otherwise. \(\theta\) can be set to 0.5 or determined by calibration experiments.
- \textbf{Soft judgment (gradient)}: directly use \(P(\text{entailment})\) as a continuous score on \([0,1]\).

\textbf{Why is this not rubric?} The key distinction: rubric requires scoring an abstract dimension like ``character development'' across the entire work --- this is a subjective aesthetic judgment that NLI models cannot reliably perform. QUIET requires judging ``whether this text explains the reason for the brain abnormality'' --- this is a logical reasoning judgment about specific content, within the core capability range of NLI models. Constraint satisfaction checking is cognitively closer to reading comprehension than literary review.

\subsection{Tier 2: Surprise}\label{tier-2-surprise}

Given that the constraint is satisfied, this measures the degree of creativity/surprise in the blank filling.

\textbf{Theoretical goal}: the surprise measure assesses how much the evaluated model's filling deviates from readers' naive expectations.

\textbf{Operationalization}: in our experiments, we use \textbf{embedding space cosine distance} as a computable implementation of surprise:

\[\text{surprise}(x_k) = 1 - \cos\left(\mathbf{e}(x_k),\; \bar{\mathbf{e}}_k\right)\]

where \(\mathbf{e}(x_k)\) is the unit vector obtained by mapping blank filling \(x_k\) through a multilingual embedding model (this paper uses Google \texttt{text-multilingual-embedding-002}), and \(\bar{\mathbf{e}}_k\) is the centroid (unitized) of the embedding vectors of all 12 evaluated models' fillings for the same blank. The core idea: when a model's filling is far from the consensus center of other models in embedding space, that model is considered to have made a non-trivial choice for that blank.

\textbf{Relationship between this operationalization and theoretical surprisal}: true token-level surprisal \(-\log P_{\text{ref}}(x_k)\) requires the probability output of a reference language model, measuring how rare the text is under the pretraining corpus distribution. In a creative benchmark setting, this implementation has a known pitfall: token-level surprisal is dominated by surface text frequency, not conceptual leaps in the creative sense. The embedding distance method avoids this pitfall to some extent --- it measures semantic deviation from consensus, not pretraining rarity of word sequences. However, this method also has limitations: when all evaluated models converge on the same mediocre answer, that cluster also becomes the centroid, and truly creative answers far from the centroid score high --- this is intuitively appropriate. Conversely, when a model gives an outlandish but semantically novel answer, its surprise will also be high --- this edge case is filtered by the constraint satisfaction score.

\textbf{Semantic clarification about logprobs}: in existing benchmarks such as HellaSwag, logprobs are used for quality judgment --- the higher the probability a model assigns to an option, the better that option is judged to be. The embedding distance component in QUIET uses this in exactly the opposite direction: the greater the embedding distance (the more it deviates from the consensus center) = the more it deviates from conventional expectations = the more valuable it is (given constraint satisfaction). The reference embedding centroid here plays not the role of judge, but of a conventionality baseline --- representing the consensus center of all evaluated models. Therefore, this component does not need to possess high-order creative aesthetic capability; it only needs to faithfully reflect the embedding distribution of common narrative patterns. Constraint satisfaction (§4.1) is responsible for judging correctness; embedding distance (§4.2) is responsible for measuring how far from consensus --- the former is a logical reasoning task, the latter is geometric distance computation, and neither requires the scoring model itself to have creative aesthetic judgment.

\subsection{Tier 3: Composite Scoring --- ``Calibrated Surprise''}\label{tier-3-composite-scoring-calibrated-surprise}

\[\text{score}(x_k) = \text{satisfy}(x_k, c_k) \times \left(1 + \lambda \cdot \text{surprise}(x_k)\right), \quad \lambda = 1.0\]

This formula encodes the semantic: ``accuracy is the foundation, surprise is the modifier'' --- when surprise is zero, score = satisfy (mediocre filling retains the base score); when surprise is maximum, score = 2·satisfy (brilliant filling gets double reward); when satisfy is zero, score = 0 (inaccurate answers score zero regardless of surprise). In our experiments, we use λ = 1.0 by default; §5.5 reports sensitivity results where ranking Spearman ≥ 0.839 holds across λ ∈ \{0.5, 1.0, 1.5\}.

Three outcome patterns of this formula:

{\def\LTcaptype{none} 
\begin{longtable}[]{@{}
  >{\raggedright\arraybackslash}p{(\linewidth - 8\tabcolsep) * \real{0.2000}}
  >{\raggedright\arraybackslash}p{(\linewidth - 8\tabcolsep) * \real{0.2000}}
  >{\raggedright\arraybackslash}p{(\linewidth - 8\tabcolsep) * \real{0.2000}}
  >{\raggedright\arraybackslash}p{(\linewidth - 8\tabcolsep) * \real{0.2000}}
  >{\raggedright\arraybackslash}p{(\linewidth - 8\tabcolsep) * \real{0.2000}}@{}}
\toprule\noalign{}
\begin{minipage}[b]{\linewidth}\raggedright
Case
\end{minipage} & \begin{minipage}[b]{\linewidth}\raggedright
satisfy
\end{minipage} & \begin{minipage}[b]{\linewidth}\raggedright
surprise
\end{minipage} & \begin{minipage}[b]{\linewidth}\raggedright
Composite Score
\end{minipage} & \begin{minipage}[b]{\linewidth}\raggedright
Meaning
\end{minipage} \\
\midrule\noalign{}
\endhead
\bottomrule\noalign{}
\endlastfoot
Constraint not satisfied & ≈ 0 & any & ≈ 0 & Inaccurate = zero score \\
Constraint satisfied but mediocre & ≈ 1 & low & low & Accurate but unsurprising \\
Constraint satisfied and brilliant & ≈ 1 & high & \textbf{high} & \textbf{Calibrated surprise ✓} \\
\end{longtable}
}

This formula is the direct operationalization of ``Calibrated Surprise'': in \(I(X;Y) = H(X) - H(X|Y)\), when \(H(X|Y) \approx 0\) (constraint satisfied → satisfy ≈ 1) and \(H(X)\) is high (surprising → surprise high), \(I(X;Y)\) takes its maximum value.

\subsection{Cascade Consistency Scoring}\label{cascade-consistency-scoring}

For blank pairs with a dependency edge \((k \to j)\) in the cascade constraint graph, cascade consistency is additionally computed:

\[\text{cascade}(x_k, x_j) = P_{\text{NLI}}(\text{entailment} \mid x_k,\; \text{cascade\_criterion}(k, j))\]

where \(\text{cascade\_criterion}(k, j)\) is the cascade consistency judgment criterion preset by the task author (e.g., ``whether the content of Blank 3 is logically consistent with the brain abnormality described in Blank 1''). This score checks whether the evaluated party maintains cascade logical consistency across multiple blanks.

\subsection{Total Score}\label{total-score}

\[\text{Total} = \sum_{g=1}^{7} \text{satisfy}_g \times (1 + \lambda \cdot \text{surprise}_g), \quad \lambda = 1.0\]

The summation ranges over all 7 blank groups, with a theoretical maximum of 7 × 2.0 = 14.0. The cascade consistency term (§4.4) is not included in the current experimental version and is left for future work.

\textbf{Composite formula family used in sensitivity analysis (§5.5 Finding 4).} Beyond the main Scheme C above, three additional functional forms are evaluated as ablation:

\begin{itemize}
\tightlist
\item
  \textbf{Scheme A} (weighted power product): \(\text{satisfy}^{1.5} \times \text{surprise}^{1.0}\)
\item
  \textbf{Scheme B} (harmonic mean): \(2 \cdot \text{satisfy} \cdot \text{surprise} / (\text{satisfy} + \text{surprise})\), zero if either side is zero
\item
  \textbf{Scheme C} (main, accuracy as base, surprise as amplifier): \(\text{satisfy} \times (1 + \lambda \cdot \text{surprise})\)
\item
  \textbf{Scheme D} (geometric mean): \(\sqrt{\text{satisfy} \cdot \text{surprise}}\)
\end{itemize}

Scheme C is adopted as the main composite because it preserves the asymmetric ``accuracy gates surprise'' structure of the calibrated-surprise framework (Zou \& Xu, 2026a): a filling that fails the constraint receives zero regardless of surprise. Schemes A, B, D appear only in the robustness check in §5.5 Finding 4.

\textbf{Aggregation method}: for each blank group, the satisfy score uses the arithmetic mean of all constraint scores within the group (soft mean aggregation); group-level surprise is the arithmetic mean of all blank-level normalized surprise within the group. Group scores are substituted into the §4.3 composite formula, then summed across 7 groups to obtain the model's total score.

\section{Experiments}\label{experiments}

\subsection{Experimental Setup}\label{experimental-setup}

\textbf{Evaluated models.} 12 commercial LLMs spanning multiple generations and vendors were selected:

{\def\LTcaptype{none} 
\begin{longtable}[]{@{}lll@{}}
\toprule\noalign{}
Model & Vendor & Category \\
\midrule\noalign{}
\endhead
\bottomrule\noalign{}
\endlastfoot
Claude-Opus-4.6 & Anthropic & Non-reasoning \\
Claude-Kaiku-4.5 & Anthropic & Non-reasoning \\
Gemini-3.1-Pro & Google & Non-reasoning \\
Gemini-2.5-pro & Google & Non-reasoning \\
GPT-5.4-mini & OpenAI & Non-reasoning \\
GPT-3.5-turbo & OpenAI & Non-reasoning \\
Hy3-preview-thinking & Tencent & Reasoning \\
ERNIE-5.1-thinking & Baidu & Reasoning \\
GLM-5.1-thinking & Zhipu & Reasoning \\
Qwen-3.6-Plus-thinking & Alibaba & Reasoning \\
Qwen-Max & Alibaba & Non-reasoning \\
Grok-4.1-fast & xAI & Non-reasoning \\
\end{longtable}
}

Note: ``Category'' only distinguishes whether the model uses an explicit reasoning chain (thinking mode), and makes no implicit ordering claim about vendor generation or parameter scale. GPT-3.5-turbo, released in 2022 as a previous-generation baseline model, serves as the lower-bound generational comparison. The remaining models are commercial flagship or main commercial models released in 2025--2026; parameter scales and training data details across vendors are mostly undisclosed, and this paper makes no cross-model positioning claims.

\textbf{Testing method.} All models received the same standardized prompt via chat interface (``Please read the following mystery story, replace all blanks with your filled-in content, and output a complete, directly readable story''), completing the task in one session in a new conversation.

\textbf{Satisfy scoring system.} Constraint satisfaction judgment uses a \textbf{three-model ensemble} (judge ensemble):
- claude-opus-4-6 (Anthropic)
- gpt-5.4-mini (OpenAI)
- gemini-3-pro-preview (Google)

Each constraint is scored independently by all three judges using a \textbf{6-point scale (integers 0--5)}, normalized to \([0,1]\) and averaged arithmetically as the final score for that constraint. The prompt explicitly provides 6 anchor points (0=completely absent or conflicts with constraint; 1=only sparse traces; 2=halfway there; 3=largely satisfied but with minor flaws; 4=fully satisfied without flaws; 5=fully satisfied and elegantly executed), with a \textbf{knockout clause}: if the constraint contains explicit ``required'' language (``must,'' ``cannot,'' ``not allowed,'' mandatory conditions connected by ``and,'' etc.) and the text violates them, the score is capped at ≤ 1 regardless of how well other aspects are written. Each blank group's satisfy score uses the arithmetic mean of all constraint scores within the group (soft mean aggregation), rather than the stricter bucket principle min(constraint\_means). Comparison of these two aggregation methods and robustness analysis are in §5.5.

\textbf{Scoring scale granularity ablation}: to rule out the possibility that ``low α stems from too coarse a scoring scale,'' this paper additionally re-ran the judge scoring on the same answer set using a coarse 3-tier 0/0.5/1.0 scoring system as an ablation control. The main results (§5.2--§5.4, all tables) report fine-grained 6-point + knockout results; the ablation control and diagnostic findings are discussed in §5.5 Finding 4 and §6.3 Limitations.

\textbf{Judge agreement}: Krippendorff α across all 20 constraints × 12 models = 240 scoring points is \textbf{0.270} (ordinal scale, using interval disagreement measure, computed from raw 0--5 integers). This is lower than the typical judge agreement in factual QA tasks (α ≈ 0.5--0.7), indicating that logical judgment of creative constraints itself has significant interpretation space. \textbf{Key diagnostic observation}: in the scoring scale granularity ablation experiment, coarse 3-tier 0/0.5/1.0 scoring gives α = 0.273 and fine-grained 6-point + knockout gives α = 0.270 --- consistent within statistical noise, indicating that \textbf{scoring scale granularity is not the bottleneck for judge agreement}; the low agreement stems from systematic judgment-criterion drift (model-level cognitive drift) between judge LLM models, not a scale issue eliminable through prompt engineering. This limitation's interpretation and improvement path are further discussed in §6.3.

\subsection{Constraint Satisfaction Scoring Results}\label{constraint-satisfaction-scoring-results}

\textbf{Table 1: Constraint satisfaction scores for 12 models × 7 blank groups (6-point scale + knockout, normalized to \([0,1]\), soft mean aggregation. Sorted by QUIET total score; see Table 3)}

{\def\LTcaptype{none} 
\begin{longtable}[]{@{}
  >{\centering\arraybackslash}p{(\linewidth - 18\tabcolsep) * \real{0.1026}}
  >{\raggedright\arraybackslash}p{(\linewidth - 18\tabcolsep) * \real{0.0769}}
  >{\centering\arraybackslash}p{(\linewidth - 18\tabcolsep) * \real{0.1026}}
  >{\centering\arraybackslash}p{(\linewidth - 18\tabcolsep) * \real{0.1026}}
  >{\centering\arraybackslash}p{(\linewidth - 18\tabcolsep) * \real{0.1026}}
  >{\centering\arraybackslash}p{(\linewidth - 18\tabcolsep) * \real{0.1026}}
  >{\centering\arraybackslash}p{(\linewidth - 18\tabcolsep) * \real{0.1026}}
  >{\centering\arraybackslash}p{(\linewidth - 18\tabcolsep) * \real{0.1026}}
  >{\centering\arraybackslash}p{(\linewidth - 18\tabcolsep) * \real{0.1026}}
  >{\centering\arraybackslash}p{(\linewidth - 18\tabcolsep) * \real{0.1026}}@{}}
\toprule\noalign{}
\begin{minipage}[b]{\linewidth}\centering
Rank
\end{minipage} & \begin{minipage}[b]{\linewidth}\raggedright
Model
\end{minipage} & \begin{minipage}[b]{\linewidth}\centering
G1
\end{minipage} & \begin{minipage}[b]{\linewidth}\centering
G2
\end{minipage} & \begin{minipage}[b]{\linewidth}\centering
G3
\end{minipage} & \begin{minipage}[b]{\linewidth}\centering
G4
\end{minipage} & \begin{minipage}[b]{\linewidth}\centering
G5
\end{minipage} & \begin{minipage}[b]{\linewidth}\centering
G6
\end{minipage} & \begin{minipage}[b]{\linewidth}\centering
G7
\end{minipage} & \begin{minipage}[b]{\linewidth}\centering
Mean
\end{minipage} \\
\midrule\noalign{}
\endhead
\bottomrule\noalign{}
\endlastfoot
1 & Gemini-3.1-Pro & 0.92 & 0.76 & 0.63 & 0.67 & 0.67 & 0.83 & 0.87 & \textbf{0.763} \\
2 & Gemini-2.5-pro & 0.85 & 0.76 & 0.67 & 0.64 & 0.47 & 0.73 & 0.82 & \textbf{0.706} \\
3 & GLM-5.1-thinking & 0.83 & 0.69 & 0.63 & 0.64 & 0.50 & 0.67 & 0.78 & \textbf{0.677} \\
4 & Claude-Opus-4.6 & 0.88 & 0.69 & 0.73 & 0.73 & 0.60 & 0.80 & 0.87 & \textbf{0.757} \\
5 & Qwen-3.6-Plus-thinking & 0.91 & 0.62 & 0.63 & 0.51 & 0.67 & 0.73 & 0.78 & \textbf{0.693} \\
6 & Grok-4.1-fast & 0.84 & 0.69 & 0.63 & 0.60 & 0.60 & 0.70 & 0.78 & \textbf{0.691} \\
7 & GPT-5.4-mini & 0.88 & 0.71 & 0.63 & 0.60 & 0.70 & 0.80 & 0.76 & \textbf{0.726} \\
8 & Claude-Kaiku-4.5 & 0.87 & 0.78 & 0.70 & 0.71 & 0.57 & 0.47 & 0.71 & \textbf{0.686} \\
9 & Hy3-preview-thinking & 0.81 & 0.62 & 0.63 & 0.58 & 0.63 & 0.80 & 0.71 & \textbf{0.684} \\
10 & Qwen-Max & 0.80 & 0.69 & 0.63 & 0.56 & 0.50 & 0.70 & 0.67 & \textbf{0.649} \\
11 & ERNIE-5.1-thinking & 0.75 & 0.73 & 0.60 & 0.60 & 0.67 & 0.53 & 0.71 & \textbf{0.656} \\
12 & GPT-3.5-turbo & 0.75 & 0.60 & 0.50 & 0.53 & 0.40 & 0.43 & 0.56 & \textbf{0.538} \\
\end{longtable}
}

\subsection{Surprise Scoring Results}\label{surprise-scoring-results}

\textbf{Scoring method}: surprise is implemented via embedding space cosine distance (see §4.2 for details). For each blank position, the fillings of all 12 evaluated models are mapped to unit vectors via Google \texttt{text-multilingual-embedding-002}, and each model's filling vector's cosine distance to the centroid of all 12 models is computed, with per-blank normalization to obtain surprise scores. Specifically, for the 12 raw cosine distance values \(d_m\) at each blank position, normalization is done by ``dividing by the maximum value at that blank position'': \(\widetilde{d}_m = d_m / \max_{m'} d_{m'}\), compressing scores to \([0,1]\) while preserving the relative distance ratios among the 12 models (implementation details in \texttt{compute\_surprise.py}). Each blank group's surprise score is the arithmetic mean of all blank-level normalized surprise within the group.

\textbf{Table 2: Surprise scores for 12 models × 7 blank groups (embedding cosine distance, normalized)}

{\def\LTcaptype{none} 
\begin{longtable}[]{@{}
  >{\centering\arraybackslash}p{(\linewidth - 18\tabcolsep) * \real{0.1026}}
  >{\raggedright\arraybackslash}p{(\linewidth - 18\tabcolsep) * \real{0.0769}}
  >{\centering\arraybackslash}p{(\linewidth - 18\tabcolsep) * \real{0.1026}}
  >{\centering\arraybackslash}p{(\linewidth - 18\tabcolsep) * \real{0.1026}}
  >{\centering\arraybackslash}p{(\linewidth - 18\tabcolsep) * \real{0.1026}}
  >{\centering\arraybackslash}p{(\linewidth - 18\tabcolsep) * \real{0.1026}}
  >{\centering\arraybackslash}p{(\linewidth - 18\tabcolsep) * \real{0.1026}}
  >{\centering\arraybackslash}p{(\linewidth - 18\tabcolsep) * \real{0.1026}}
  >{\centering\arraybackslash}p{(\linewidth - 18\tabcolsep) * \real{0.1026}}
  >{\centering\arraybackslash}p{(\linewidth - 18\tabcolsep) * \real{0.1026}}@{}}
\toprule\noalign{}
\begin{minipage}[b]{\linewidth}\centering
Rank
\end{minipage} & \begin{minipage}[b]{\linewidth}\raggedright
Model
\end{minipage} & \begin{minipage}[b]{\linewidth}\centering
G1
\end{minipage} & \begin{minipage}[b]{\linewidth}\centering
G2
\end{minipage} & \begin{minipage}[b]{\linewidth}\centering
G3
\end{minipage} & \begin{minipage}[b]{\linewidth}\centering
G4
\end{minipage} & \begin{minipage}[b]{\linewidth}\centering
G5
\end{minipage} & \begin{minipage}[b]{\linewidth}\centering
G6
\end{minipage} & \begin{minipage}[b]{\linewidth}\centering
G7
\end{minipage} & \begin{minipage}[b]{\linewidth}\centering
Mean
\end{minipage} \\
\midrule\noalign{}
\endhead
\bottomrule\noalign{}
\endlastfoot
1 & GLM-5.1-thinking & 0.751 & 0.847 & 0.860 & 0.939 & 0.633 & 0.661 & 0.703 & \textbf{0.771} \\
2 & GPT-3.5-turbo & 0.577 & 0.735 & 0.804 & 0.780 & 0.836 & 0.677 & 0.629 & \textbf{0.720} \\
3 & Gemini-2.5-pro & 0.824 & 0.823 & 0.727 & 0.746 & 0.556 & 0.482 & 0.720 & \textbf{0.697} \\
4 & Qwen-3.6-Plus-thinking & 0.670 & 0.598 & 0.632 & 0.793 & 0.607 & 0.879 & 0.641 & \textbf{0.689} \\
5 & Claude-Kaiku-4.5 & 0.668 & 0.500 & 0.657 & 0.752 & 0.697 & 0.909 & 0.624 & \textbf{0.687} \\
6 & Grok-4.1-fast & 0.676 & 0.716 & 0.779 & 0.814 & 0.567 & 0.632 & 0.572 & \textbf{0.679} \\
7 & Qwen-Max & 0.624 & 0.565 & 0.626 & 0.736 & 0.647 & 0.640 & 0.917 & \textbf{0.679} \\
8 & Gemini-3.1-Pro & 0.470 & 0.650 & 0.867 & 0.800 & 0.607 & 0.579 & 0.526 & \textbf{0.643} \\
9 & ERNIE-5.1-thinking & 0.435 & 0.628 & 0.667 & 0.830 & 0.556 & 0.812 & 0.538 & \textbf{0.638} \\
10 & Hy3-preview-thinking & 0.615 & 0.628 & 0.663 & 0.743 & 0.508 & 0.471 & 0.618 & \textbf{0.607} \\
11 & Claude-Opus-4.6 & 0.446 & 0.651 & 0.588 & 0.748 & 0.643 & 0.526 & 0.521 & \textbf{0.589} \\
12 & GPT-5.4-mini & 0.661 & 0.498 & 0.615 & 0.710 & 0.573 & 0.457 & 0.577 & \textbf{0.584} \\
\end{longtable}
}

Under the embedding distance implementation, surprise discrimination is relatively balanced across groups; GLM-5.1-thinking (mean 0.771) and Gemini-2.5-pro (mean 0.697) lead on surprise, while Claude-Opus-4.6 (0.589) and GPT-5.4-mini (0.584) have lower surprise scores --- in contrast to their higher constraint satisfaction scores (0.757 and 0.726 respectively) (see §5.5 Finding 3).

\subsection{Composite Scoring Results}\label{composite-scoring-results}

\textbf{Composite scoring formula (Scheme C, λ=1.0):}

\[\text{QUIET\_score}_g = \text{satisfy}_g \times (1 + \lambda \cdot \text{surprise}_g), \quad \lambda = 1.0\]

\[\text{Total} = \sum_{g=1}^{7} \text{QUIET\_score}_g\]

A mediocre filling that satisfies constraints retains the base score (satisfy × 1.0 = satisfy); a brilliant filling that satisfies constraints earns up to double reward (satisfy × 2.0); a filling that fails constraints scores 0. Theoretical maximum: 14.0.

\textbf{Table 3: QUIET composite score rankings for 12 models (6-point satisfy + soft mean + Scheme C, λ=1.0)}

{\def\LTcaptype{none} 
\begin{longtable}[]{@{}clccc@{}}
\toprule\noalign{}
Rank & Model & satisfy mean & surprise mean & \textbf{QUIET Total} \\
\midrule\noalign{}
\endhead
\bottomrule\noalign{}
\endlastfoot
1 & Gemini-3.1-Pro & 0.763 & 0.643 & \textbf{8.69} \\
2 & Gemini-2.5-pro & 0.706 & 0.697 & \textbf{8.44} \\
3 & GLM-5.1-thinking & 0.677 & 0.771 & \textbf{8.40} \\
4 & Claude-Opus-4.6 & 0.757 & 0.589 & \textbf{8.38} \\
5 & Qwen-3.6-Plus-thinking & 0.693 & 0.689 & \textbf{8.18} \\
6 & Grok-4.1-fast & 0.691 & 0.679 & \textbf{8.11} \\
7 & GPT-5.4-mini & 0.726 & 0.584 & \textbf{8.03} \\
8 & Claude-Kaiku-4.5 & 0.686 & 0.687 & \textbf{8.03} \\
9 & Hy3-preview-thinking & 0.684 & 0.607 & \textbf{7.67} \\
10 & Qwen-Max & 0.649 & 0.679 & \textbf{7.62} \\
11 & ERNIE-5.1-thinking & 0.656 & 0.638 & \textbf{7.46} \\
12 & GPT-3.5-turbo & 0.538 & 0.720 & \textbf{6.44} \\
\end{longtable}
}

Theoretical maximum: 7 groups × 2.0 = 14.0 (satisfy score of 1.0 × maximum surprise bonus of 2.0).

\begin{figure}[H]
  \centering
  \includegraphics[width=0.95\linewidth]{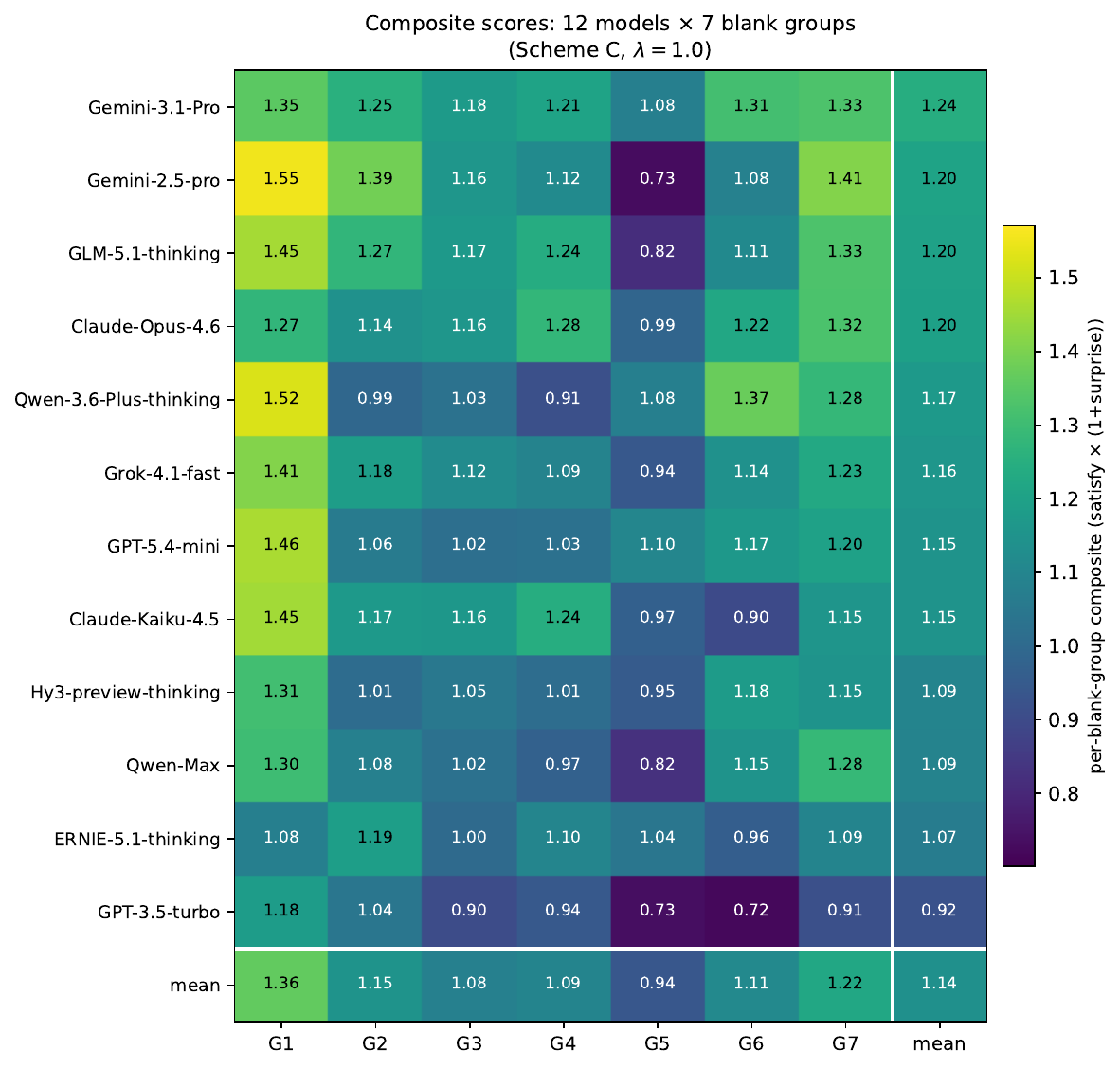}
  \caption{Composite QUIET score heatmap across 12 models \(\times\) 7 blank
  groups (Scheme C, \(\lambda{=}1.0\); cell value \(=\)
  \(\text{satisfy} \times (1+\text{surprise})\), range \([0,2]\)). Rows are
  sorted by QUIET Total (Table 3). The rightmost column and bottom row report
  per-model and per-group means; the bottom-right cell is the grand mean. The
  heatmap exposes the within-cluster gap noted in Finding 2: most modern
  models cluster in the 1.0--1.4 band, while GPT-3.5-turbo's bottom row stays
  consistently below 1.0.}
  \label{fig:heatmap}
\end{figure}

\subsection{Analysis}\label{analysis}

\textbf{Finding 1: Generational gap between GPT-3.5-turbo and the modern model cluster.} Among the 12 models, only GPT-3.5-turbo (6.44) scores significantly below the rest (7.46--8.69). This confirms that the benchmark has clear discriminative power in separating previous-generation baseline models from modern flagship and mid-tier models. GPT-3.5-turbo's mean satisfy (0.538) is notably lower than the 0.649--0.763 range for other models.

\textbf{Finding 2: Limited discrimination within the modern model cluster.} The 9 models ranked 3rd through 11th have QUIET totals spanning 7.46--8.40, a gap of 0.94 on a 14-point scale. For comparison, the same range under coarse 3-tier scoring spans 0.67 --- the 6-point scale + knockout clause provides roughly +40\% improvement in spreading apart the middle tier, but absolute discrimination remains constrained. Satisfy means cluster at 0.649--0.757 and surprise means at 0.584--0.771. The remaining discrimination limitations and improvement paths are discussed in §6.3.

\textbf{Finding 3: Ranking divergence between satisfy and surprise --- confirming the independence of the two dimensions.} Different models show distinct strength profiles across the two dimensions:
- Gemini-3.1-Pro: satisfy=0.763, surprise=0.643 → leads on satisfy, mid-range surprise (total 8.69, rank 1)
- Claude-Opus-4.6: satisfy=0.757, surprise=0.589 → high accuracy but conservative (total 8.38, rank 4)
- GLM-5.1-thinking: satisfy=0.677, surprise=0.771 → leads on surprise, weaker satisfy (total 8.40, rank 3)
- GPT-3.5-turbo: satisfy=0.538, surprise=0.720 → decent surprise but inaccurate (total 6.44, rank 12)

This distribution confirms that the surprise dimension carries independent information not replaceable by satisfy. If ranked on satisfy alone, GLM-5.1-thinking (satisfy rank 9th) would be severely undervalued, yet it rises to 3rd place in the composite score. Conversely, Claude-Opus-4.6, which ranks 2nd on satisfy, drops to 4th in the total score due to its lower surprise.

\begin{figure}[H]
  \centering
  \includegraphics[width=0.85\linewidth]{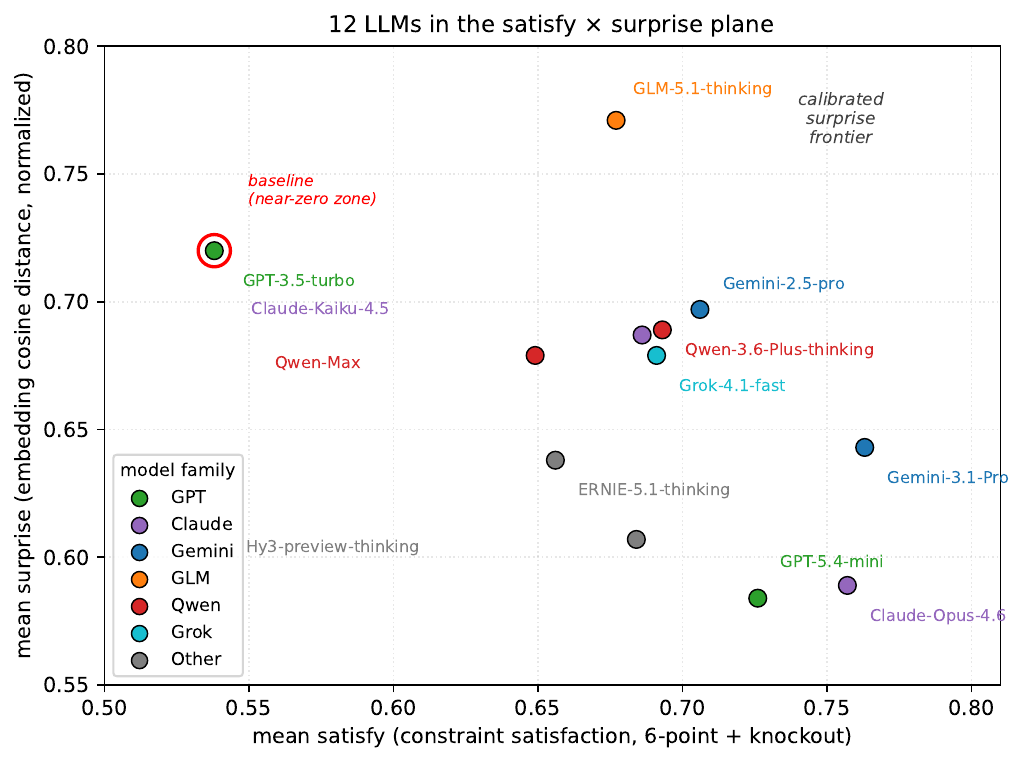}
  \caption{Two-dimensional view of the 12 evaluated models on the
  satisfy--surprise plane (per-model means across the 7 blank groups; data
  from Tables 1 and 2). Colors encode model family. GPT-3.5-turbo, circled in
  red, anchors the lower-left ``inaccurate / mid-surprise'' region; the modern
  cluster spans a roughly diagonal band, illustrating that satisfy and surprise
  carry independent information --- no single dimension would reproduce the
  composite ranking in Table 3.}
  \label{fig:satisfy-surprise}
\end{figure}

\textbf{Finding 4: Sensitivity analysis --- robustness across aggregation and formula variants.} Across 14 aggregation configurations, rank Spearman correlations among model total scores show the following:

\begin{itemize}
\tightlist
\item
  Under soft mean aggregation, the five formula variants --- Scheme C (λ=0.5/1.0/1.5), Scheme A, and Scheme D --- all show pairwise rank Spearman ≥ 0.86. GPT-3.5 ranking last and the Gemini family ranking highest remain stable across all configurations.
\item
  Fixing Scheme C (λ=1.0), the rank Spearman between soft mean and hard bucket aggregation = 0.881 --- the 6-point scale + knockout clause already handles single-constraint failures at the atom level, significantly shrinking the gap between hard and soft aggregation. This paper uses soft mean as the primary result for consistency.
\item
  The rank Spearman between this paper's main result (soft mean + Scheme C, λ=1.0) and the naive hard-bucket formula (coarse 3-tier satisfy + hard-gated aggregation) = \textbf{0.93} --- demonstrating that across the joint change from coarse to fine-grained satisfy scoring scale and from hard-gated to soft Scheme C, the final ranking remains highly consistent. This paper's conclusions do not depend on scoring detail choices.
\end{itemize}

The complete robustness matrix is in appendix data table \texttt{final\_scores.spearman.csv}.

\textbf{Finding 5: Impact of single-constraint failure.} The satisfy data shows cases of severe single-constraint penalization for Claude-Kaiku-4.5 (G6), ERNIE-5.1-thinking (G6), and GPT-3.5-turbo (G5/G6). Under bucket aggregation, such failures can drag an entire group's score down to approximately 0.2--0.4; under soft mean, they affect the group average by only approximately 0.1--0.2. The knockout clause already explicitly handles ``mandatory wording violated'' hard failures at the atom level (e.g., Gemini-3.1-Pro G7 = 0.87, corresponding to one knockout trigger in Table 1), but soft flaws still depend on aggregation choice. Justification for soft mean: if a story has multiple decision points with only a few flawed, it should not intuitively be judged as a failed work --- a continuous satisfy description better matches the actual distribution of creative quality than hard constraints.

\subsection{Human Pilot Study}\label{human-pilot-study}

The design of QUIET's test set and the cascade constraint theoretical framework were not induced post hoc from LLM experiments. They were inspired by behavioral distributions observed in a human creative writing community prior to the automated scoring protocol. This section reports the setup and qualitative findings of that pilot human cloze experiment, which serve as an independent, prior, qualitative corroboration of the theoretical framework (§2) and scoring protocol (§4).

\textbf{Experimental setup.} The experiment was conducted in March 2026 by a study collaborator in a small invitation-only online community of literary writing enthusiasts. The same mystery story test (7 blank groups × 36 blanks, identical to the §5 LLM experiment) was distributed without a time limit, and participants voluntarily submitted within a few days. N = 135 participants, self-reporting sustained writing experience or strong literary reading preferences. This experiment was completed \textbf{before} the QUIET scoring protocol was designed; its sole purpose was to validate the test's answerability and the reasonableness of the creative constraints --- no automated scoring was involved.

\textbf{Spontaneous three-tier behavioral stratification.} Without any intervention from the task author, participants' submissions self-sorted into three categories:

{\def\LTcaptype{none} 
\begin{longtable}[]{@{}
  >{\raggedright\arraybackslash}p{(\linewidth - 4\tabcolsep) * \real{0.3333}}
  >{\raggedright\arraybackslash}p{(\linewidth - 4\tabcolsep) * \real{0.3333}}
  >{\raggedright\arraybackslash}p{(\linewidth - 4\tabcolsep) * \real{0.3333}}@{}}
\toprule\noalign{}
\begin{minipage}[b]{\linewidth}\raggedright
Category
\end{minipage} & \begin{minipage}[b]{\linewidth}\raggedright
Count (\%)
\end{minipage} & \begin{minipage}[b]{\linewidth}\raggedright
Behavioral Description
\end{minipage} \\
\midrule\noalign{}
\endhead
\bottomrule\noalign{}
\endlastfoot
\textbf{Mid-task dropout} & 100 (74.07\%) & Submitted partial answers then abandoned remaining blanks; self-reported: ``it gets harder as you go, my earlier choices blocked the path forward'' / ``I can't make it work'' \\
\textbf{Submitted but mediocre} & 31 (22.96\%) & Submitted complete answers satisfying all constraints, but content stayed at the most obvious default options (e.g., brain abnormality = amnesia, profession = detective) \\
\textbf{Shortlisted candidates} & 4 (2.96\%) & Submitted complete answers; multiple blanks simultaneously satisfied constraints and were genuinely surprising; all 4 required light editorial revision by the task author --- none could be used without modification \\
\end{longtable}
}

\textbf{Correspondence with theory.} This three-tier distribution provides \textbf{prior, independent, qualitative} empirical support for the theoretical framework of §2:

\begin{enumerate}
\def\labelenumi{\arabic{enumi}.}
\item
  \textbf{``Mid-task dropout'' corresponds to the extreme form of the cascade chain rule.} This phenomenon directly reflects the chain rule in §2.2: \(I(X_k; Y \mid X_{<k}) \to 0\) --- the cumulative choices in the first \(k-1\) blanks narrow the feasible solution space for the \(k\)-th blank to the empty set, and participants cannot complete subsequent blanks while maintaining accuracy. This phenomenon is entirely absent in the LLM experiment (all 12 models completed all 36 blanks), suggesting that human participants possess a ``know when to stop'' cascade perception ability that current LLMs lack.
\item
  \textbf{``Submitted but mediocre'' corresponds to the ``accurate without surprise'' quadrant in §2.1.} These participants satisfy all constraints (high satisfy) but lack surprise (low surprise) --- the empirical instantiation of the ``mediocre correctness'' pattern predicted by the CQA framework. The modern LLM cluster in our experiments (satisfy 0.65--0.76, surprise 0.58--0.77) is behaviorally similar to this quadrant.
\item
  \textbf{``Shortlisted candidates'' correspond to the ``calibrated surprise'' quadrant.} The rare combination of high satisfy ∧ high surprise, with even the shortlisted human responses requiring further editing --- this constitutes empirical evidence that high-mutual-information creative choices are intrinsically scarce (§2.1). It also suggests that the current LLM scoring ceiling (maximum 8.69/14.0) remains significantly below expert human performance.
\end{enumerate}

\textbf{Comparison with LLM results.} The three-tier human behavioral pattern yields two meaningful contrasts with the 12-model LLM scores:

\begin{itemize}
\tightlist
\item
  LLM total scores range from 6.44 to 8.69, with no model reaching the ``accurate ∧ surprising'' standard represented by the 4 shortlisted human submissions. This qualitative gap is consistent with the ceiling imposed by absolute satisfy/surprise values.
\item
  LLMs do not ``give up'' --- all models generate filled content for all 36 blanks, even when the filling clearly violates earlier constraints (see §5.5 Finding 5 for single-constraint failure cases). This behavioral difference suggests that QUIET can serve not only to evaluate creative generation quality but also to probe models' metacognitive awareness of ``unsolvable'' cascade states --- an evaluation dimension beyond the reach of single-blank benchmarks.
\end{itemize}

\begin{figure}[H]
  \centering
  \includegraphics[width=0.95\linewidth]{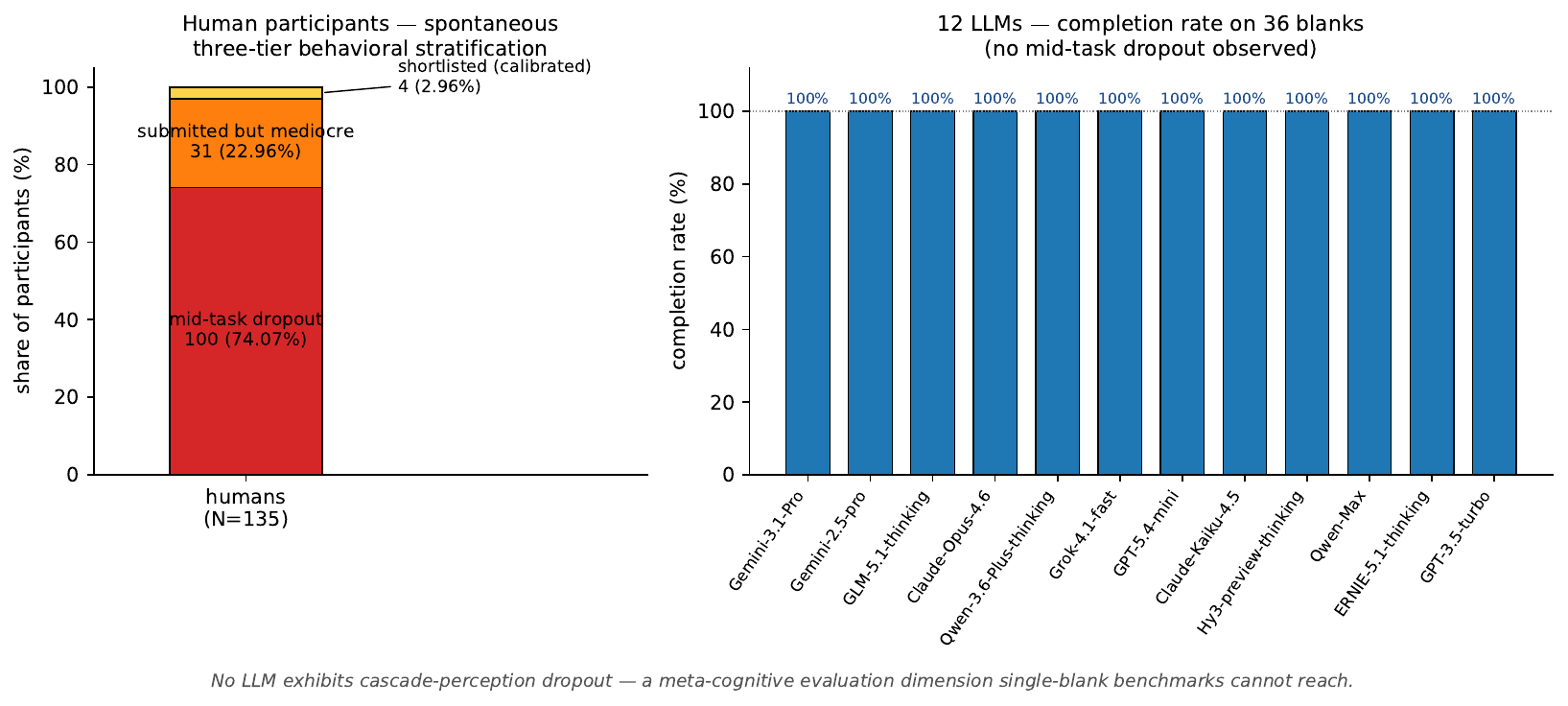}
  \caption{Completion-pattern contrast between the human pilot study (left,
  \(N{=}135\)) and the 12 evaluated LLMs (right). Humans spontaneously
  stratified into three tiers --- mid-task dropout (74.07\%),
  submitted-but-mediocre (22.96\%), and shortlisted (2.96\%) --- whereas every
  LLM completed all 36 blanks, with no observed analogue of the human
  ``give-up under accumulating constraints'' behavior. This contrast motivates
  treating cascade-state metacognition as a distinct evaluation dimension
  beyond raw generation quality.}
  \label{fig:human-vs-llm}
\end{figure}

\textbf{Note.} Due to the limited recruitment scale and the non-anonymous nature of the community, the pilot study (§5.6) did not collect demographic information, and participants' fillings were not subjected to unified quantitative scoring. This section therefore reports qualitative observations only and is not included in the main statistical results of §5.1--5.5. A fully quantitative human comparison experiment (with a larger sample across age groups and writing backgrounds) is left as future work, to be conducted jointly with the rubric-based atomic check scoring scheme proposed in §6.4.

\section{Discussion}\label{discussion}

\subsection{Structural Differences from Rubric-Based Evaluation}\label{structural-differences-from-rubric-based-evaluation}

The difference between rubric-based evaluation and QUIET is not merely a methodological preference; there are irreducible structural differences at two levels.

\textbf{Handling of dimension interaction effects.} Rubrics implicitly assume linear additivity: \(\text{Quality} = \sum_i w_i \cdot s_i\). But creative quality depends on \(I(X;\, Y_1 \cap \cdots \cap Y_k)\), where dimension interaction effects are the core mechanism driving the sharp drop in conditional entropy. Scoring dimensions independently is mathematically equivalent to discarding all interaction terms. QUIET's constraint satisfaction scoring does not split dimensions --- each blank's constraint \(c_k\) is a content requirement integrating multi-dimensional information, and the NLI model naturally accounts for intra-constraint dimension interactions when making its judgment.

\textbf{Different levels of evaluation object.} Rubric evaluation assesses an entire work's performance across dimensions --- its input is a complete text, its output is a set of dimension scores. QUIET evaluates each blank's constraint satisfaction --- its input is a piece of filled text and a constraint condition, its output is a satisfaction score. The evaluation granularity and cognitive operations are entirely different and should not be conflated.

It should be noted that this paper does not claim rubrics are useless in all contexts. For highly standardized alignment tasks, rubrics have clear scaling advantages. This paper's argument is limited to: when the evaluation target involves generation-side measurement of higher-order creative ability, QUIET's per-blank constraint satisfaction + information-theoretic scoring provides precision that rubrics cannot achieve.

The rubric critique here targets rubrics in \textbf{generation-side} use (decomposing quality into linearly weighted dimensions for training or generation alignment). The rubric-based atomic check items this paper plans to adopt in future measurement work is rubric in its \textbf{measurement-side} use --- functionally distinct from the object of critique in this section; see §6.4 for details.

\subsection{Comparison with Existing Benchmarks}\label{comparison-with-existing-benchmarks}

{\def\LTcaptype{none} 
\begin{longtable}[]{@{}
  >{\raggedright\arraybackslash}p{(\linewidth - 12\tabcolsep) * \real{0.1429}}
  >{\raggedright\arraybackslash}p{(\linewidth - 12\tabcolsep) * \real{0.1429}}
  >{\raggedright\arraybackslash}p{(\linewidth - 12\tabcolsep) * \real{0.1429}}
  >{\raggedright\arraybackslash}p{(\linewidth - 12\tabcolsep) * \real{0.1429}}
  >{\raggedright\arraybackslash}p{(\linewidth - 12\tabcolsep) * \real{0.1429}}
  >{\raggedright\arraybackslash}p{(\linewidth - 12\tabcolsep) * \real{0.1429}}
  >{\raggedright\arraybackslash}p{(\linewidth - 12\tabcolsep) * \real{0.1429}}@{}}
\toprule\noalign{}
\begin{minipage}[b]{\linewidth}\raggedright
Benchmark
\end{minipage} & \begin{minipage}[b]{\linewidth}\raggedright
Blanks
\end{minipage} & \begin{minipage}[b]{\linewidth}\raggedright
Task Form
\end{minipage} & \begin{minipage}[b]{\linewidth}\raggedright
Scoring
\end{minipage} & \begin{minipage}[b]{\linewidth}\raggedright
Cascade Constraints
\end{minipage} & \begin{minipage}[b]{\linewidth}\raggedright
Measurement Target
\end{minipage} & \begin{minipage}[b]{\linewidth}\raggedright
\(\rho\) Assumption
\end{minipage} \\
\midrule\noalign{}
\endhead
\bottomrule\noalign{}
\endlastfoot
Story Cloze (2016) & 1 & 2-choice & Accuracy & No & Recognition & Implicit \(\rho \approx 0\) \\
HellaSwag (2019) & 1 & 4-choice & logprobs & No & Recognition & Implicit \(\rho \approx 0\) \\
ARC (2018) & 1 & 4-choice & logprobs & No & Recognition & Implicit \(\rho \approx 0\) \\
MMLU (2021) & 1 & 4-choice & logprobs & No & Recognition & Implicit \(\rho \approx 0\) \\
\textbf{QUIET (this paper)} & \textbf{10--20} & \textbf{Open generation} & \textbf{NLI + embedding distance} & \textbf{Yes} & \textbf{Creative generation} & \textbf{Naturally avoided} \\
\end{longtable}
}

All existing benchmarks in the table use a single-blank + multiple-choice paradigm and share two implicit assumptions: (1) \(\rho \approx 0\) --- quality can be judged from prior context alone; (2) recognition ability ≈ creative ability --- being able to select a good answer ≈ being able to write one. QUIET breaks both assumptions: the multi-blank full-structure design naturally handles \(\rho\); open generation directly measures creative ability.

\subsection{Limitations}\label{limitations}

\begin{enumerate}
\def\labelenumi{\arabic{enumi}.}
\item
  \textbf{Limited test set scale.} The initial version has a limited number of stories and blanks; statistical power may be insufficient and expansion is needed.
\item
  \textbf{Constraint design requires expert involvement.} Writing constraint conditions requires participation from creative domain experts, which is both a methodological contribution and a scaling constraint.
\item
  \textbf{Bias in the NLI scoring model.} Automated scoring relies on the NLI model's capabilities --- if the NLI model itself has limited understanding of narrative constraints, scoring will introduce systematic bias. Subsequent work will recruit human experts to independently judge filling results and report the correlation between automated scores and human expert judgments to quantify this bias.
\item
  \textbf{Two caveats for the embedding distance centroid method.} The surprise dimension is implemented via geometric distance from the 12-model embedding centroid (§4.2), introducing two known biases. First, \textbf{reference-set dependence} --- the centroid is determined by the currently evaluated 12 models; different sets of evaluated models yield different surprise rankings. Second, \textbf{cross-generation outward deviation misread as creative surprise} --- in a model set mixing generations (this paper includes both GPT-3.5-turbo from 2022 and modern flagships from 2025--2026), a previous-generation baseline model's fillings systematically deviate from the modern model consensus center in embedding space, and the centroid method counts this generational difference as surprise. In the actual data, GPT-3.5-turbo's mean surprise of 0.720 exceeds Gemini-3.1-Pro (0.643) and Claude-Opus-4.6 (0.589), exemplifying this phenomenon. This does not affect the main conclusion of the composite score (GPT-3.5 still ranks last due to extremely low satisfy), but suggests that future work should use a more robust surprise baseline --- e.g., recomputing the centroid after excluding outliers with large generational gaps, or introducing shortlisted human responses as an absolute anchor.
\item
  \textbf{Cross-cultural applicability unvalidated.} The current test set is primarily based on Chinese narrative material.
\item
  \textbf{Multi-judge ensemble averaging obscures subtle constraints.} This paper's satisfy scoring uses a 3-model LLM judge ensemble averaged arithmetically. The ensemble improves scoring stability but introduces a known bias: when a constraint contains subtle implicit semantics (e.g., ``the prop must have archival properties''), if any one judge fails to strictly identify this and gives full marks, the ensemble mean is inflated, masking a filling that should have been scored as partially satisfied. A representative case is Gemini-3.1-Pro's use of ``medicine'' as the key prop in G7 (violating the constraint that ``props cannot be made on the spot''). Under the coarse 3-tier scoring system (ablation control), this case's satisfy was judged at 1.0; the 6-point scale + knockout clause used in this paper's main result reduced it to 0.87, indicating the knockout mechanism already partially identifies such violations. But completely eliminating masking requires the atom-level rubric path; §6.4 gives this improvement direction.
\item
  \textbf{Discrimination within the modern model cluster remains limited --- but the bottleneck has been localized through scoring scale granularity ablation.} The 9 models ranked 3rd through 11th have QUIET totals spanning 7.46--8.40 (on a 14-point scale, effective discrimination span of 0.94), an improvement over the 0.67 span under the coarse 3-tier system --- showing that the 6-point scale + knockout clause does help spread apart the middle tier. However, Krippendorff α barely changed across the two scoring scales (coarse 3-tier: 0.273 vs.~fine-grained 6-point + knockout: 0.270; in \(\alpha = 1 - D_o/D_e\), observed and expected disagreement scale together, leaving their ratio unchanged), constituting a \textbf{diagnostic negative result}: the granularity of the scoring scale is not the bottleneck for discrimination and consistency. \textbf{The real bottleneck is systematic judgment-criterion drift between judge models} --- the disagreement among three judges on the same constraint (within-item) is close to their random disagreement across different constraints (between-item), suggesting the judges lack a shared judgment anchor. This finding corrects a naive intuition that ``rubric decomposition is the primary improvement path'': breaking holistic LLM-as-Judge into finer atom-level checks has theoretical value (§6.4), but without simultaneously introducing \textbf{human-anchored calibration} (pre-annotating K samples per constraint at N levels with human experts as reference points, then injecting these as in-context examples into judge prompts), simply increasing atom count risks distributing the same model-level cognitive drift across more atoms without substantially improving α. The engineering design of the next scoring protocol and the implementation path for human-anchored calibration are in §6.4.
\end{enumerate}

\subsection{Future Work}\label{future-work}

\textbf{From current version to scale.} The current version is a proof of concept. Subsequent work includes: expanding the test set (more stories, more blanks), cross-language extension (Chinese-English bilingual test set), and exploration of automated constraint generation.

\textbf{Closed-loop validation with CQA fine-tuning.} QUIET can serve as an independent validation tool for the fine-tuning effects reported in the CQA paper (Zou \& Xu, 2026c): score differences on QUIET before and after CQA fine-tuning for the same model constitute external evidence of fine-tuning effectiveness.

\textbf{Cross-domain transfer.} QUIET's design principle --- multi-blank cascade constraints + information-theoretic automated scoring --- is not limited to narrative creation. Any domain requiring judgment under multi-step cascade constraints (e.g., legal reasoning, medical diagnostic reasoning, design evaluation) can theoretically adopt a similar benchmark design paradigm.

\textbf{Next scoring protocol: human-anchored calibration + rubric-based satisfy.} The scoring scale granularity ablation experiment reported in §5.1 and §6.3 identifies the priority direction for improvement: after expanding the scoring scale from coarse 3-tier to fine-grained 6-point + knockout, Krippendorff α remained stable at 0.27 (see §6.3 item 7), indicating that scale granularity is not the bottleneck for judge consistency --- the real bottleneck is systematic judgment-criterion drift between judge models. The next scoring protocol therefore needs two simultaneous improvements: (a) \textbf{human-anchored calibration} --- pre-annotate K cross-level samples per constraint with N human experts as a reference distribution, inject these as in-context examples into judge prompts, enabling the three LLM judges to score against the same reference distribution; (b) \textbf{rubric-based atomic check items} --- decompose each constraint into 2--4 binary atomic questions (e.g., ``can the prop remain available throughout the story timeline,'' ``can the prop store/record information''), have LLM judges adjudicate each item, then synthesize a total score by rule. The team has completed the full design of the atomic check items --- 56 atoms total, of which 37 are marked as required (★) --- and specified 6 controlled experiments (E1--E6) to verify the compound effects of both improvements on α, discrimination, and single-constraint failure robustness. This improvement path has established methodology in educational assessment (Messick, 1994).

\textbf{Important clarification}: the rubric use this paper criticizes in §1.1 and §6.1 is rubric in its \textbf{generation-side} application --- decomposing quality into weighted dimensions for training alignment or generation guidance, producing checklist-style creative mediocrity. The rubric-based scoring this paper's future work proposes is rubric in its \textbf{measurement-side} application --- having scorers consistently judge already-existing samples against atomic check items to improve scoring reliability. These two uses are functionally entirely distinct in the measurement literature (cf.~Messick, 1994 on construct representativeness and consequential validity): the former is a generation constraint, the latter is a post-hoc evaluation tool.

\section{Conclusion}\label{conclusion}

This paper introduces QUIET (Quality Understanding via Interlocked Evaluation Testing), a diagnostic benchmark for LLM creative ability based on multi-blank cascade story cloze and information-theoretic automated scoring.

QUIET's design contains four core innovations. First, 10--20 interlocked blanks are placed within a complete-structure story, with cascade constraint dependencies between blanks --- each blank's filling constrains the feasible solution space for subsequent blanks --- forcing the test subject to make local creative decisions from a global perspective. Second, open generation rather than a multiple-choice recognition paradigm is used, directly upgrading the evaluation target from creative judgment to creative generation. Third, an automated scoring protocol based on constraint satisfaction × surprise is designed, directly operationalizing the ``calibrated surprise'' information-theoretic framework --- constraint satisfaction is automatically judged by an NLI model (objective logical inference, not subjective aesthetics), and surprise is computed as the embedding cosine distance from a multi-model consensus centroid (§4.2); the two are multiplied to obtain a composite score without human grading. Fourth, retrospective sensitivity \(\rho(X_t)\) is formalized and it is shown how the multi-blank full-structure design naturally avoids the retrospective evaluation dilemma faced by single-blank paradigms --- this theoretical contribution simultaneously reveals the \(\rho \approx 0\) scope assumption that existing mainstream benchmarks (HellaSwag, Story Cloze, etc.) have long implicitly assumed but never explicitly stated.

Pilot experiments on 12 models corroborate the theoretical predictions. On the constraint satisfaction dimension, the three mid-chain groups G3--G5 (where cumulative cascade constraints are densest) show significantly lower cross-model mean satisfy than G1 (opening) and G7 (closing), with G5 having the lowest cross-model mean satisfy (approximately 0.58), validating the increasing discrimination pattern of the cascade constraint accumulation effect. On the surprise dimension (implemented via embedding cosine distance), different models show varying degrees of semantic deviation in their creative choices. The surprise and satisfy dimensions show imperfect correlation, and the two constitute independent information. The composite score reveals creative path differences masked by satisfy alone: certain models (e.g., GLM-5.1-thinking) receive significant surprise bonuses that move them from a mid-low satisfy position (9th) into the composite score front ranks (3rd). We also transparently report a known limitation of the LLM-as-Judge ensemble scoring system --- Krippendorff α ≈ 0.27 moderate inter-rater reliability --- and use a scoring scale granularity ablation experiment (coarse 3-tier vs.~fine-grained 6-point + knockout) as a diagnostic control to localize the source of this limitation as systematic judgment-criterion drift between judge models rather than scoring scale granularity, with a corresponding next-version improvement plan of human-anchored calibration + rubric-based atomic check items proposed in §6.4.

\textbf{Human behavioral evidence independent of the automated scoring protocol.} It is worth emphasizing that the theoretical framework of this work was not induced retrospectively from LLM scoring results. Before the QUIET scoring protocol was designed, the same test was distributed to 135 literary writing enthusiasts in an invitation-only community. Without any scoring prompts or human intervention, participants spontaneously stratified into three tiers: ``mid-task dropout,'' ``submitted but mediocre,'' and ``shortlisted candidates'' (§5.6), corresponding respectively to the extreme form of the chain rule, the ``accurate without surprise'' quadrant, and the ``calibrated surprise'' quadrant. This human behavioral distribution --- which \textbf{emerged prior to framework construction} --- provides independent prior corroboration of this paper's theoretical framework. It also highlights a systematic absence in LLMs of ``know when to stop'' cascade perception: all 12 evaluated models completed all 36 blank fills, with none exhibiting the dropout behavior human participants showed when facing saturated cascade constraints. This contrast not only validates the ecological validity of QUIET's evaluation dimensions but also reveals a new evaluation dimension that single-blank benchmarks have long been unable to reach: a model's metacognitive awareness of ``unsolvable'' cascade states.

QUIET provides the first benchmark framework based on an information-theoretic product structure for reproducible, scalable quantitative evaluation of LLM creative ability, filling a long-standing gap between open generation testing and objective automated scoring.

\section*{References}
\addcontentsline{toc}{section}{References}

\begin{itemize}
\tightlist
\item
  Shannon, C. E. (1948). A Mathematical Theory of Communication. \emph{The Bell System Technical Journal, 27}(3), 379--423.
\item
  Mostafazadeh, N., et al.~(2016). A Corpus and Cloze Evaluation for Deeper Understanding of Commonsense Stories. \emph{Proceedings of NAACL-HLT}.
\item
  Zellers, R., et al.~(2019). HellaSwag: Can a Machine Really Finish Your Sentence? \emph{Proceedings of ACL}.
\item
  Clark, P., et al.~(2018). Think You Have Solved Question Answering? Try ARC, the AI2 Reasoning Challenge. \emph{arXiv:1803.05457}.
\item
  Hendrycks, D., et al.~(2021). Measuring Massive Multitask Language Understanding. \emph{Proceedings of ICLR}.
\item
  Ye, S., et al.~(2024). FLASK: Fine-grained Language Model Evaluation based on Alignment Skill Sets. \emph{Proceedings of ICLR (Spotlight)}.
\item
  Krippendorff, K. (2018). \emph{Content Analysis: An Introduction to Its Methodology} (4th ed.). SAGE Publications.
\item
  Messick, S. (1994). The Interplay of Evidence and Consequences in the Validation of Performance Assessments. \emph{Educational Researcher, 23}(2), 13--23.
\item
  van Dijk, T. A., \& Kintsch, W. (1983). \emph{Strategies of Discourse Comprehension}. Academic Press.
\item
  Sternberg, M. (2003). Universals of Narrative and Their Cognitivist Fortunes (I). \emph{Poetics Today, 24}(2), 297--395.
\item
  Chatman, S. (1978). \emph{Story and Discourse: Narrative Structure in Fiction and Film}. Cornell University Press.
\item
  Graesser, A. C., Singer, M., \& Trabasso, T. (1994). Constructing Inferences During Narrative Text Comprehension. \emph{Psychological Review, 101}(3), 371--395.
\item
  Ely, J., Frankel, A., \& Kamenica, E. (2015). Suspense and Surprise. \emph{Journal of Political Economy, 123}(1), 215--260.
\item
  Wilmot, D., \& Keller, F. (2020). Modelling Suspense in Short Stories as Uncertainty Reduction over Neural Representation. \emph{Proceedings of ACL}.
\item
  Zou, B., \& Xu, C. (2026a). Calibrated Surprise: An Information-Theoretic Account of Creative Quality. arXiv:2604.26269.
\item
  Zou, B., \& Xu, C. (2026b). BC Protocol: Structured Dual-Expert Dialogue for Eliciting High-Quality Chain-of-Thought Post-Training Data. arXiv preprint (forthcoming, 2026).
\item
  Zou, B., \& Xu, C. (2026c). Creative Quality Alignment: Expert Tacit Knowledge Transfer via Chain-of-Thought Fine-Tuning. arXiv preprint (forthcoming, 2026).
\end{itemize}

\end{document}